\documentclass[lettersize,journal]{IEEEtran}

\usepackage{amsmath,amsfonts}
\usepackage{algorithmic}
\usepackage{array}
\usepackage[caption=false,font=normalsize,labelfont=sf,textfont=sf]{subfig}
\usepackage{textcomp}
\usepackage{stfloats}
\usepackage{url}
\usepackage{verbatim}
\usepackage{graphicx}
\def\BibTeX{{\rm B\kern-.05em{\sc i\kern-.025em b}\kern-.08em
    T\kern-.1667em\lower.7ex\hbox{E}\kern-.125emX}}
\usepackage{balance}
\usepackage[colorlinks=true, urlcolor=blue]{hyperref}        
\usepackage{booktabs}
\usepackage{multirow}

\usepackage{xcolor}  
\usepackage{framed}  

\definecolor{highlightgray}{gray}{0.95}

\begin{document}

\title{IMPACT: A Generic Semantic Loss for Multimodal Medical Image Registration}
\author{Valentin Boussot, Cédric Hémon, Jean-Claude Nunes, Jason Dowling, Simon Rouzé, Caroline Lafond, Anaïs Barateau, Jean-Louis Dillenseger 
\thanks{V. Boussot, C. Hémon, J-C. Nunes, C. Lafond, A. Barateau and J-L. Dillenseger are with Univ. Rennes, CLCC Eugene Marquis, INSERM, LTSI - UMR 1099, F-35000 Rennes, France.\\
J. Dowling is with CSIRO Australian e-Health Research Centre, Herston, Queensland, Australia.\\
S. Rouzé is with CHU Rennes, Department of Cardio-Thoracic and Vascular Surgery, F-35000 Rennes, France.}}

\maketitle

\begin{abstract}
Image registration is fundamental in medical imaging, enabling precise alignment of anatomical structures for diagnosis, treatment planning, image-guided interventions, and longitudinal monitoring. This work introduces IMPACT (Image Metric with Pretrained model-Agnostic Comparison for Transmodality registration), a novel similarity metric designed for robust multimodal image registration. Rather than relying on raw intensities, handcrafted descriptors, or task-specific training, IMPACT defines a semantic similarity measure based on the comparison of deep features extracted from large-scale pretrained segmentation models. By leveraging representations from models such as TotalSegmentator, Segment Anything (SAM), and other foundation networks, IMPACT provides a task-agnostic, training-free solution that generalizes across imaging modalities. These features, originally trained for segmentation, offer strong spatial correspondence and semantic alignment capabilities, making them naturally suited for registration. The method integrates seamlessly into both algorithmic (Elastix) and learning-based (VoxelMorph) frameworks, leveraging the strengths of each. IMPACT was evaluated on five challenging 3D registration tasks involving thoracic CT/CBCT and pelvic MR/CT datasets. Quantitative metrics, including Target Registration Error and Dice Similarity Coefficient, demonstrated consistent improvements in anatomical alignment over baseline methods. Qualitative analyses further highlighted the robustness of the proposed metric in the presence of noise, artifacts, and modality variations. With its versatility, efficiency, and strong performance across diverse tasks, IMPACT offers a powerful solution for advancing multimodal image registration in both clinical and research settings.
\end{abstract}

\begin{IEEEkeywords}
Image Registration, Semantic similarity measure, Foundation models
\end{IEEEkeywords}

\section{Introduction}

Image registration is a fundamental technique in medical imaging that establishes precise anatomical correspondences between multiple scans through spatial transformations. Establishing this correspondence is essential for various clinical and research applications, including contour propagation, disease diagnosis, treatment planning, image-guided treatment and longitudinal monitoring. It enables the integration of information from multiple and complementary imaging modalities, such as Magnetic Resonance Imaging (MRI), Computed Tomography (CT), Cone Beam CT (CBCT), and Positron Emission Tomography (PET), to provide a more complete understanding of anatomical structures and functional processes. Moreover, accurate registration allows the tracking of anatomical changes over time, for instance, in monitoring tumor progression or assessing treatment response \cite{chen_recent_2022,viergever_survey_2016}.

In unsupervised intensity-based registration, alignment is achieved by optimizing a deformation model through the minimization of a cost function that combines a similarity measure and a regularization term. The similarity measure quantifies the alignment quality between the fixed and moving images, while the regularization term enforces a smooth, anatomically plausible spatial transformation \cite{dalca_unsupervised_2019,de_vos_deep_2019,chen2022transmorph}.

Similarity measures play an essential role in image registration, directly influencing its effectiveness. However, accurately quantifying similarity is challenging due to the complex, non-linear relationships between multimodal images, which stem from differences in imaging modalities, noise, and acquisition artifacts. Traditional similarity measures such as mean squared error (MSE), normalized cross-correlation (NCC) \cite{studholme1996automated}, and mutual information (MI) \cite{viola1997alignment} rely on intensity-based comparisons, assuming that well-aligned images exhibit consistent or statistically related intensities. Although effective in some contexts, these metrics often fail in multimodal scenarios where intensity relationships are not preserved. 
To overcome these limitations, various handcrafted features have been introduced to incorporate local structural information beyond raw intensity values. Notable examples include the Modality Independent Neighborhood Descriptor (MIND) \cite{heinrich_mind_2012} and the Self-Similarity Context (SSC) \cite{heinrich2013towards}, both designed to capture spatial patterns that remain stable across modalities. These descriptors demonstrated greater robustness in a variety of registration tasks. However, they remain limited by their sensitivity to noise, their reliance on fixed parameter tuning, and their inability to encode high-level anatomical semantics. Since they rely exclusively on local intensity self-similarity, they can fail in cases of large deformations or anatomical variations, leading to suboptimal alignments \cite{gao_monomodal_2008,penney1998comparison,hering2022learn2reg,shiraishi2018image}.

Recent deep learning–based similarity measures address these limitations by replacing handcrafted descriptors with hierarchical representations learned from data \cite{haskins_learning_2019,ronchetti_disa_2023,czolbe_semantic_2023,chen_survey_2025,liu2021same,song2024dino,siebert2024convexadam}. Instead of relying on handcrafted features, these methods leverage deep neural networks to extract multi-scale feature embeddings that capture both structural consistency and semantic context. This shift marks a transition from purely intensity-based registration toward approaches grounded in high-level semantic understanding of image content.

This transition aligns with recent advances for image synthesis and segmentation. Deep feature-based losses, such as the perceptual loss using VGG embeddings \cite{johnson2016perceptual}, have replaced traditional pixel-wise differences by comparing high-level feature representations rather than raw intensities. By leveraging hierarchical features extracted from pretrained networks, these methods better capture structural and textural correspondences, ensuring that generated images maintain perceptual fidelity while remaining robust to intensity variations and noise. 
Similarly, in medical image segmentation, comparing deep feature maps has proved more effective than traditional geometrical measures.
This feature-based approach allows models to align segmentation outputs with ground truth annotations in a way that better preserves anatomical structures, ensuring sharper boundaries and improved structural coherence \cite{pihlgren2023systematic}.

Although these perceptual similarity metrics have shown remarkable success in image synthesis and segmentation, their potential for image registration remains largely unexplored. In particular, no existing approach has leveraged deep feature representations from large-scale models to construct a robust, task-agnostic similarity measure for multimodal registration.

Building on the successes of perceptual loss and handcrafted features, we propose a new feature-based similarity measure for image registration that fully exploits large-scale pretrained models. While previous deep learning-based approaches have explored feature extraction for image registration, they do not explicitly formulate similarity measures as differentiable loss functions \cite{liu2021same,siebert2024convexadam}. Among these, only one utilizes a generic pretrained model trained on a large-scale dataset without requiring additional task-specific training \cite{song2024dino}, and only one prior work has proposed a dedicated similarity loss, though it still relies on task-specific training \cite{czolbe_semantic_2023}. 

However, these methods share fundamental limitations: they require task-specific training and operate on feature maps instead of images, leading to suboptimal convergence and spatial inconsistencies, particularly in models with large receptive fields like transformers or deep CNNs. By relying solely on extracted feature maps, these approaches overlook the spatial sensitivity of deep features to transformations. Moreover, they rely on rigid feature extraction frameworks that limit adaptability across various imaging modalities and registration tasks.

In contrast, our method directly leverages the Jacobian of the pretrained model to enable a more principled and spatially coherent optimization process. Furthermore, our approach exploits high-level representations of large-scale generic segmentation models, reassigning them as robust feature extractors for similarity estimation. Moreover, our framework remains highly flexible by formulating similarity estimation as a differentiable loss function, allowing seamless integration into a wide range of registration techniques, including algorithmic and deep learning-based frameworks.

Furthermore, with the increasing availability of generic pretrained models trained on large-scale datasets, leveraging these architectures has become particularly relevant. Models trained on extensive and diverse datasets to segment multiple structures within images learn hierarchical feature representations. These representations capture rich spatial and contextual information, increasing the robustness of similarity measures while eliminating the need for extensive task-specific training. 

In this work, we present IMPACT (Image Metric with Pretrained model-Agnostic Comparison for Transmodality Registration), a novel generic deep learning-based similarity metric that removes the need for training task-specific models. IMPACT exploits ready-made segmentation models (TotalSegmentator, SAM2.1) as universal 2D/3D feature extractors. By effectively comparing semantic features derived from these large-scale architectures, the proposed method achieves improved alignment in both algorithmic and deep learning–based registration frameworks. 

Our contributions are threefold:
\begin{itemize}
    \item We propose a novel similarity measure that leverages features from large-scale pretrained models such as TotalSegmentator \cite{wasserthal_totalsegmentator_2023}, SAM2.1 \cite{kirillov2023segment} for multimodal medical image registration
    \item We establish that these generic features serve as robust and generalizable descriptors for multimodal registration tasks, eliminating the need for large annotated datasets and task-specific training while maintaining high accuracy.  
    \item We evaluate the efficiency and versatility of our method by integrating it into both Elastix \cite{klein_elastix_2010} (algorithmic registration) and VoxelMorph \cite{balakrishnan_voxelmorph_2019} (deep learning-based registration), leveraging the strengths of each framework, and showing consistent improvements across multiple anatomical regions and imaging modalities.
\end{itemize}

IMPACT has been extensively evaluated across five challenging multimodal registration tasks, integrating seamlessly into both algorithmic (Elastix) and deep learning-based (VoxelMorph) frameworks. The evaluation encompassed diverse imaging modalities (CBCT, CT, MRI) and anatomical regions (abdomen, pelvis, thorax), and included benchmarks from multiple public challenges. The results demonstrate consistent improvements in alignment accuracy across varying image qualities, acquisition protocols, and anatomical variability, highlighting IMPACT's robustness and generalizability in real-world clinical settings.

\section{Background and Related Work}
\subsection{Image Registration: An Optimization Problem}
In unsupervised intensity-based image registration, the objective is to estimate a spatial transformation $T_\theta$: $\Omega_F \rightarrow \Omega_M$ that aligns a moving image $I_M$ : $\Omega_M \rightarrow \mathbb{R}$ with a fixed image $I_F$ : $\Omega_F \rightarrow \mathbb{R}$. Both images are $D$-dimensional, defined over their respective spatial domains, $\Omega_F \subset \mathbb{R}^D and \Omega_M \subset \mathbb{R}^D$. The application of the transformation to the moving image $I_M$ is expressed through function composition: $I_M \circ T_\theta$. The goal is to find the optimal transformation by minimizing a cost function $\mathcal{C}$ \eqref{eq:1}, which depends on a similarity metric $\mathcal{S}$ and a $\gamma$-weighted regularization term $\mathcal{P}$. As the central component of the registration process, $\mathcal{S}$ quantifies the alignment quality between the transformed moving image and the fixed image. Its design is crucial, as it controls the robustness, accuracy, and generalization capability of the registration framework, making it the core focus of this study. To complement this, the regularization term $\mathcal{P}(T_\theta)$ enforces spatial smoothness by penalizing gradients or curvature (bending energy) of the transformation, ensuring a trade-off between deformation flexibility and anatomical plausibility \cite{reithmeir2024model}.

This leads to the following optimization problem:
\begin{gather}
\hat{\theta} = \underset{\theta}{\arg\min } \text{ } \mathcal{C}(I_F, I_M \circ T_\theta ) \label{eq:optiproblem}\\
\text{with } \mathcal{C}(I_F, I_M \circ T_\theta ) = -\mathcal{S}(I_F, I_M \circ T_\theta) + \gamma \mathcal{P}(T_\theta) \label{eq:1}
\end{gather}

Solving this optimization problem requires either iterative optimization for each image pair or learning-based approaches that generalize over datasets. These lead to two main paradigms in image registration: algorithmic approaches and deep learning-based methods.

\subsection{Registration Framework}

Algorithmic approaches iteratively optimize $\mathcal{C}$ for each image pair, typically relying on gradient-based optimization and explicit deformation models. Among these, B-splines \cite{Rueckert1999} provide smooth, regularized transformations by parameterizing the deformation field, while Demons algorithms \cite{thirion1998image} iteratively estimate displacement fields based on optical flow principles. To enforce topology preservation, diffeomorphic models such as LDDMM \cite{beg2005computing} and SyN \cite{avants2008symmetric} have been introduced. However, optimizing high-dimensional deformation models can be challenging. To improve convergence and capture both global and local deformations, hierarchical multiresolution strategies are commonly employed \cite{Rueckert1999}. By progressively refining the transformation from coarse to fine scales, these methods enhance robustness and prevent local minima in complex deformation spaces. These methods are widely implemented in Elastix \cite{klein_elastix_2010} and ANTs \cite{avants2009advanced}, the two most commonly used open-source frameworks for medical image registration. Both are built on the ITK (Insight Toolkit) framework and provide extensive tools for intensity-based registration. Discrete optimization has emerged as a powerful alternative to continuous methods, offering improved handling of large deformations and reduced sensitivity to local minima. Deeds \cite{heinrich2013mrf} adopts an MRF framework with densely sampled displacements and efficient regularization via a minimum spanning tree. CorrField \cite{heinrich2015estimating} extends this idea by introducing sparse keypoints, symmetry constraints, and parts-based regularization to achieve high accuracy even in cases of strong respiratory motion. Recent efforts combine discrete, continuous, and hybrid strategies to further improve registration robustness and accuracy \cite{siebert2024convexadam}.

In contrast, deep learning methods take advantage of CNNs \cite{balakrishnan_voxelmorph_2019} or transformer-based architectures \cite{chen2022transmorph} to directly predict $T_\theta$ from a pair of images, eliminating the need for iterative optimization during inference. Recent advancements in these architectures primarily aim to enhance their generalization ability, ensuring robust performance on unseen test data across diverse anatomical structures and imaging modalities. However, training these models remains an iterative algorithmic process, where the loss function $\mathcal{C}$ is optimized over multiple iterations using gradient-based methods such as Adam. This process updates the network weights to ensure that the learned deformation field accurately models the transformations observed in the anatomical structures of the training data.

The coexistence of these two paradigms reflects their complementary strengths. Algorithmic methods provide robustness, generalizability, and reliable performance across datasets, making them well-suited for applications where data is scarce. Furthermore, some operate in physical coordinates, making them independent of image size and voxel resolution. This enhances their effectiveness, particularly in scenarios where the fields of view of the images to be registered differ significantly. 
However, their computational cost poses a challenge for real-time applications. In contrast, deep learning methods excel in speed and accuracy for task-specific applications, particularly when large, high-quality annotated datasets are available. Although these methods can achieve high accuracy, their performance highly depends on the quality and diversity of the training dataset. Furthermore, they are sensitive to domain shifts and may require fine-tuning or retraining when applied to new datasets. However, acquiring such datasets is often challenging due to variations in imaging protocols and the need for manual annotations. The choice between these approaches depends on the application requirements, balancing computational resources, data availability, and the nature of the data \cite{Jena2024}.

\subsection{Multimodal Similarity Metrics for Image Registration}

A major challenge in image registration is handling complex multimodal scenarios, where variations in intensity, noise, and acquisition artifacts introduce significant discrepancies between the images to be aligned. 
These differences can lead to misalignment when using traditional similarity metrics such as MSE and NCC, making the design of a robust similarity metric $\mathcal{S}$ essential for accurate registration. While these methods assume consistent intensity relationships in monomodal settings, this assumption breaks down in multimodal scenarios, requiring alternative approaches to account for modality-specific variations.

One of the earliest and most widely used approaches to multimodal registration relies on measuring statistical dependence between intensity distributions. MI \cite{viola1997alignment,wells1996multi} and its extensions, such as Normalized MI (NMI) \cite{studholme1999overlap}, estimate the shared information between images, making them effective for aligning images with complex intensity relationships. However, MI-based metrics can be sensitive to noise and prone to local optima due to inaccurate histogram estimation. To mitigate these limitations, gradient-based refinements such as Gradient-Intensity MI \cite{pluim2000image} integrate spatial gradient information, enhancing alignment consistency and increasing robustness to noise.

Beyond statistical dependencies, multimodal similarity metrics have evolved into three main approaches. Modality gap reduction strategies seek to minimize differences between imaging modalities, either by explicitly learning a shared representation space or by enforcing robustness to modality variations. Handcrafted descriptors extract structural features that remain stable across modalities, improving robustness to intensity variations. Deep-learned representations leverage neural networks to extract high-level semantic features, enabling robust alignment across diverse imaging modalities.

The following sections provide a detailed analysis of these approaches, highlighting their strengths and limitations in multimodal image registration.

\subsubsection{Modality Gap Reduction Strategies for Similarity Learning}

Modality gap reduction strategies aim to bridge differences between imaging modalities by enhancing similarity in image space or feature space. Some methods transform images into a shared representation, while others enforce robustness to modality variations through adversarial learning or contrast-invariant representations.

Han et al. \cite{han2022joint} propose JSR, a framework that jointly synthesizes CT images from MR and CBCT while optimizing deformation fields for multimodal alignment.
Hémon et al. \cite{hemon2024indirect} follow a sequential approach, first synthesizing target-modality images with GANs, then registering them using traditional similarity metrics. While effective, this method requires extensive training data and may introduce artifacts.
Yan et al. \cite{yan2018} introduce AIRNet, which uses adversarial learning to distinguish well-aligned from misaligned pairs, guiding the registration network without explicitly transforming images. However, adversarial training is often unstable and requires careful tuning.
Hoffmann et al. \cite{hoffmann2021synthmorph} propose SynthMorph, a contrast-invariant learning method that removes dependency on real training data by generating synthetic images with extreme contrast and deformation variations. Rather than transforming images, SynthMorph trains networks to ignore contrast differences, enabling robust registration across unseen modalities.

While these approaches improve multimodal registration, they introduce challenges. Domain adaptation and adversarial methods require large datasets and may struggle with unseen modalities, whereas contrast-invariant learning reduces dataset dependency but may not fully capture modality-specific structures.

\subsubsection{Hand-Crafted Feature-Based Similarity Metrics}

To move beyond raw intensity comparisons, structural descriptors have been introduced to capture local spatial patterns that remain stable across modalities. Normalized Gradient Fields (NGF) \cite{haber2006intensity} align images based on edge structures rather than absolute intensities, making them more robust to modality changes. Similarly, MIND \cite{heinrich_mind_2012} and SSC \cite{heinrich2013towards} extract local self-similar features, improving robustness to intensity variations. While these methods outperform statistical metrics in complex cases, they still rely on manually designed descriptors, limiting their ability to capture high-level semantic correspondences.

\subsubsection{Learned Feature-Based Similarity Metric}
The evolution of similarity metrics in medical image registration has shifted towards deep learning-based feature extraction, enabling more robust and adaptive similarity measures. Several methods have been proposed to learn feature representations that capture anatomical correspondences across modalities.

Czolbe et al. \cite{czolbe_semantic_2023} introduce a semantic similarity metric that aligns images based on learned, task-specific features rather than intensity values, improving robustness to noise and modality variations.
SAME (Self-supervised Anatomical Embeddings) \cite{liu2021same} is a monomodal registration pipeline that uses self-supervised embeddings to learn anatomical correspondences without requiring manual annotations. 
DINO-Reg \cite{song2024dino} leverages self-supervised, transformer-based features from DINOv2 for image registration, highlighting the potential of foundation models in medical image alignment.
ConvexAdam \cite{siebert2024convexadam} leverages handcrafted descriptors (MIND) or learned deep features (nnU-Net, when annotations are available) within an adaptive dual-optimization framework.

While these methods improve registration robustness, they have inherent limitations.
Many depend on task-specific training, making them highly dataset-dependent and less generalizable. Additionally, because they operate on feature maps rather than raw images, they can suffer from suboptimal convergence and spatial inconsistencies, particularly in models with large receptive fields, such as transformers or deep CNNs, as observed in DINO-Reg \cite{song2024dino}. By relying exclusively on extracted feature representations, these methods overlook the spatial sensitivity of deep features to transformations, potentially causing alignment errors. Additionally, they impose rigid feature extraction frameworks, limiting adaptability across modalities, and do not leverage the advantages of algorithmic registration techniques.

Unlike previous approaches that rely on handcrafted descriptors or task-specific deep feature learning, our method introduces a novel paradigm by using large-scale segmentation models as generic feature extractors for multimodal image registration. Instead of designing similarity metrics from scratch, we directly leverage pretrained segmentation networks, such as SAM and TotalSegmentator to provide robust, modality-agnostic feature representations for defining a semantic similarity metric.

Recent foundation segmentation models have transformed medical image analysis by providing modality-agnostic spatial representations that generalize across diverse imaging modalities. Unlike traditional CNNs, these transformer-based architectures capture long-range dependencies and global context, enabling accurate anatomical segmentation and feature-driven alignment without task-specific training.

Models like Segment Anything Model (SAM) \cite{kirillov2023segment} and MedSAM \cite{ma2024segment} extend prompt-based segmentation to medical imaging, leveraging large-scale heterogeneous datasets to enhance generalization. Similarly, TotalSegmentator \cite{wasserthal_totalsegmentator_2023} and STU-Net \cite{huang2023stu} unify segmentation across multiple modalities, reinforcing the shift from intensity-based registration to deep feature-driven alignment. These networks offer a promising avenue for improving multimodal image registration by leveraging robust anatomical representations rather than intensity similarities.

By leveraging these pretrained “foundation” segmentation models (SAM, MedSAM, TotalSegmentator, STU-Net, etc.), image registration can be guided by structural similarity rather than raw intensity values. The challenge remains in efficiently integrating these extracted feature maps into optimization frameworks, ensuring alignment is driven by semantic correspondences while maintaining computational efficiency and adaptability.

\section{Method}
\label{sec:method}
A similarity metric based on feature-driven alignment between images is proposed. Features from fixed and moving images are extracted using a selected pretrained model and compared using a selected feature distance measure. The next section introduces this feature-based metric in the context of both registration paradigms: deep learning-based methods, such as VoxelMorph \cite{balakrishnan_voxelmorph_2019}, and algorithmic approaches implemented in the Elastix framework \cite{klein_elastix_2010}.
 
\subsection{Semantic Similarity Metric Based on Pretrained Model for Algorithmic Registration}

The fundamental principles of algorithmic registration are outlined within the \textit{Elastix} framework, a widely used open-source tool in biomedical imaging that is built upon the ITK library \cite{mccormick2014itk}. \textit{Elastix} offers a comprehensive set of transformation models and robust optimization techniques, supporting various similarity metrics, including MSE, NCC, and NMI that are well-suited for both mono and multimodal applications. However, it does not support the MIND or any semantic similarity measures, motivating the development of a novel metric. This paper presents the full architecture of the registration framework, encompassing not only the proposed semantic similarity metric, but also the iterative optimization process, deformation models, and multi-resolution schemes. An advanced strategy was employed using weakly supervised registration with anatomical masks, allowing the optimization to focus on clinically relevant regions. Lastly, this work provides a comprehensive description of the similarity measure implementation, supporting two distinct operational modes (Jacobian and Static) and four distance metrics $\mathcal{D}$: L1, L2, NCC, and cosine similarity, to quantify similarity between extracted feature representations.

\subsubsection{Adaptive Stochastic Gradient Descent (ASGD)}
A fundamental aspect of the framework is its optimization strategy, which iteratively refines the transformation parameters. The optimization problem (Eq.\ref{eq:1}) is typically addressed using a first-order derivative-based iterative approach :
\begin{equation}
\theta_{i+1} = \theta_i-a_i \frac{\partial \mathcal{C}(I_F, I_M \circ T_\theta)}{\partial \theta}
\end{equation}
where $a_i$ denotes the step size or gain factor. Among the available strategies, Adaptive Stochastic Gradient Descent (ASGD) \cite{Klein2008} is one of the most commonly used due to its efficiency and robustness in medical image registration. ASGD improves computational efficiency by estimating gradients on a randomly selected subset of voxels rather than the full image domain \cite{bollapragada2018adaptive,qiao2019efficient}. This stochastic sampling significantly reduces computation time while maintaining a representative cost function. Additionally, ASGD employs an adaptive stepsize, dynamically adjusting the learning rate to stabilize updates. Unlike deterministic optimization methods, ASGD does not include a built-in convergence criterion, as the stochastic nature of gradient estimation leads to fluctuations in the cost function, making standard stopping conditions unreliable. As a result, our experiments are conducted for a fixed number of iterations to ensure consistent comparisons across different configurations.

\subsubsection{Deformation Models}
The experiments conducted with \textit{Elastix} use a free-form deformation model based on B-spline interpolation \cite{Rueckert1999}. The transformation $T_\theta$ is represented as a sparse function with a limited number of parameters. B-splines are particularly well-suited for capturing smooth and continuous deformations while maintaining computational efficiency, making them a robust choice for image registration tasks \cite{Rueckert1999}.




\subsubsection{Multiresolution Strategy}
To address the high computational complexity of optimizing transformations across the entire image domain, a multiresolution strategy is employed \cite{klein_elastix_2010}. The registration begins at a coarse resolution, where large-scale deformations are estimated, and progressively refines the alignment at higher resolutions as finer details are introduced. This hierarchical approach simplifies the optimization landscape, mitigates the risk of local minima, and enables efficient handling of both global and local transformations.

\subsubsection{Weakly Supervised Approach using Masks}
In many medical imaging applications, partial annotations or region-specific masks are available. These can be effectively used within a weakly supervised registration framework to guide the alignment toward clinically relevant structures. Instead of using the entire image domain for optimization, the \textit{Elastix} framework allows restricting the registration process to regions defined by these masks. This approach focuses the optimization on clinically relevant areas, such as organs or pathological regions, while ignoring irrelevant or noisy regions in the image. Additionally, this strategy is particularly useful for capturing complex anatomical interactions, such as sliding motions between adjacent organs \cite{Wu2008,Vandemeulebroucke2012}, by limiting the optimization to the ROI where such motions are most relevant. This ensures more accurate and anatomically meaningful registration results. 

Technically, the cost function is computed using only voxels within the mask domain \(\Omega_F^\text{mask} \subset \Omega_F\) of the fixed image and the mask domain \(\Omega_M^\text{mask} \subset \Omega_M\) of the moving image, reducing the complexity and improving the robustness of the registration. The transformation \(T_\theta\) is then optimized to align the fixed and moving images only within the masked regions.

\subsubsection{IMPACT : Jacobian mode}

\begin{figure*}[h!]
\centering
    \includegraphics[width=0.8\textwidth]{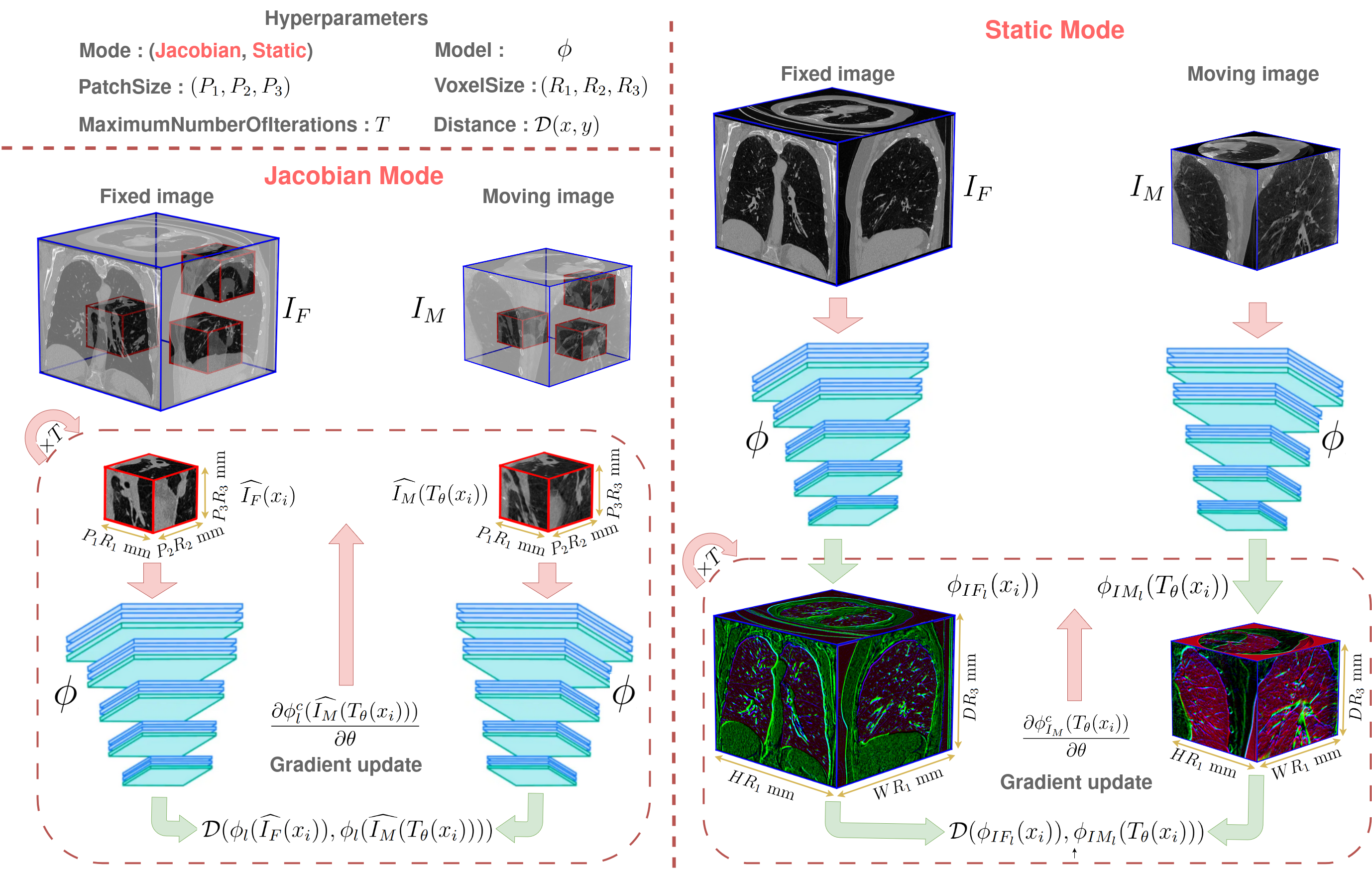}
    \caption{Schematic overview of the proposed feature-based similarity metric the algorithmic registration framework in the 3D case. Features are extracted from the fixed and moving images using a shared pretrained encoder $\phi$. The similarity is computed between features of the fixed image $\phi(I_F(x_i))$ and the transformed moving image $\phi(I_M(T_\theta(x_i)))$ using a distance function $\mathcal{D}$. Two modes are supported: Jacobian Mode, where local patches are sampled and gradients are backpropagated through the feature extractor, and Static Mode, where features are precomputed over the entire image and reused throughout the optimization. Gradients are used to iteratively update the transformation $T$, aligning the moving image to the fixed image.}
    \label{fig:method}
\end{figure*}

Fig. \ref{fig:method} provides a schematic overview of the Jacobian and static mode workflow, illustrating how patches are sampled from the fixed and moving images, passed through the shared feature extractor, and then compared to produce gradients for updating the transformation parameters.

For each iteration, $N$ patches of dimension $D$ are extracted uniformly at random. Each patch has a size $P = (P_1, P_2, \dots, P_D)$, with $P \in \mathbb{N}^D$, corresponding to the receptive field at a deeper layer. For each patch extracted from the fixed image, a corresponding patch is sampled from the moving image using the current spatial transformation. Based on these patches, the similarity measure is defined as follows:
\begin{gather}\label{eq:costJacobian}
\mathcal{S}(I_F, I_M \circ T_\theta ) = \\
\frac{1}{NC} \sum_{l=1}^L \gamma_l \sum_{i=1}^N \mathcal{D} \Big( \phi_l(\widehat{I_F}(x_i)), \phi_l(\widehat{I_M}(T_\theta(x_i))) \Big)\nonumber
\end{gather}

where $\widehat{I_F}(x_i) : \Omega_F \rightarrow \mathbb{R}^{P_1 \times P_2 \times \dots \times P_D}$ and $\widehat{I_M}(x_i) : \Omega_M \rightarrow \mathbb{R}^{P_1 \times P_2 \times \dots \times P_D}$ denote the patches extracted from the fixed and moving images, centered at the coordinate $x_i$. 

These patches are uniformly sampled such that all positions within each patch lie entirely within their respective mask regions. The extracted patches are then resampled to a resolution of \( R \in \mathbb{R}^D \) mm using third-order B-spline interpolation over the image function \( I \).

The function $\phi_l : \mathbb{R}^{P_1 \times P_2 \times \dots \times P_D} \rightarrow \mathbb{R}^C$ extracts $C$ feature representations from the patch at layer $l$, while $\gamma_l$ represents the weight assigned to that layer. The function $ \mathcal{D}(x, y)$ quantifies the similarity between two feature representations.

The focus is placed exclusively on first-order gradient-based optimization methods, specifically employing gradient descent to iteratively minimize the loss function by adjusting the transformation parameters, $\theta$. The gradient of the loss function with respect to $\theta$ is expressed as:

\begin{gather}
\label{eq:gradJacobian}
\frac{\partial \mathcal{S}(I_F, I_M \circ T_\theta)}{\partial \theta} = \\
\frac{2}{NC} \sum_{l = 1}^L \gamma_l \sum_{i = 1}^N \sum_{c = 1}^C 
\left. \frac{\partial \mathcal{D}(f, m)}{\partial m} \right|_{\begin{aligned} 
    f &= \phi_l^c(\hat{I}_F(x_i)) \\ 
    m &= \phi_l^c(\hat{I}_M(T_\theta(x_i))) \nonumber
\end{aligned}}\\
\times \frac{\partial \phi^c_l(\widehat{I_M}(T_\theta(x_i)))}{\partial \theta} \nonumber
\end{gather}

Using the chain rule, the derivative of the $c$-th feature with respect to the transformation parameters $\theta$ can be expressed as:  
\begin{equation}
\label{eq9}
\frac{\partial \phi_l^c(\widehat{I_M}(T_\theta(x_i)))}{\partial \theta} = \nabla_{\widehat{I_M}} \phi^c_l \cdot \nabla_x \widehat{I_M} \cdot \frac{\partial T_\theta(x_i)}{\partial \theta}
\end{equation}
The equation (Eq.\ref{eq9}) highlights the interplay between three components:

\begin{itemize}
    \item \textbf{Gradient of the feature extractor} ($ \nabla_{\widehat{I_M}} \phi_l^c \in \mathbb{R}^{P_1 \times P_2 \times \dots \times P_D}$):  
    This term captures how the $c$-th feature, extracted by $\phi$, responds to changes in the intensity of the image patch $\widehat{I_M}$.  

    \item \textbf{Spatial gradient of the moving image} ($\nabla_x \widehat{I_M} \in \mathbb{R}^{(P_1 \times P_2 \times \dots \times P_D) \times D}$):  
    This term measures how $\widehat{I_M}$ varies with respect to spatial coordinates $x$. For smoother results, $\nabla_x \widehat{I_M}$ is computed using third-order B-spline interpolation, which ensures stability and continuity.  

    \item \textbf{Local Jacobian of the transformation} ($\frac{\partial T_\theta(x_i)}{\partial \theta} \in \mathbb{R}^{D \times |\theta|}$):  
    This represents how the transformation $T_\theta$ deforms spatial coordinates in response to changes in its parameters $\theta$.  
\end{itemize}

By integrating these components, the resulting gradient $\frac{\partial \phi^c(\widehat{I_M}(T_\theta(x_i)))}{\partial \theta} \in \mathbb{R}^{|\theta|}$ reflects the contribution of the feature extractor $\phi$, whose gradients with respect to the moving image $I_M$ are efficiently computed via backpropagation. Notably, this computation is independent of the specific deformation model considered, as the Jacobian $\frac{\partial T_\theta(x_i)}{\partial \theta}$ encapsulates the transformation influence on the spatial coordinates. This decoupling ensures that the approach is flexible and compatible with a wide range of transformation models.

\subsubsection{IMPACT : Static mode}
The time required to compute the Jacobian of the feature extraction function $\phi$ increases with the model’s receptive field, making backpropagation through deep networks computationally intensive. While these gradients are very useful for efficiently minimizing the loss function in a few iterations, they become impractical for models with large receptive fields. However, they are not strictly necessary when the extracted features maintain spatial alignment with the input images and when the spatial sensitivity of deep features to geometric transformations is negligible. In such cases, full backpropagation can be avoided without significantly affecting alignment performance. For this reason, a slightly simplified version, referred to as Static mode, was developed for loss function computation. This approach bypasses gradient calculation by precomputing the features and directly comparing them.

In static mode, the feature maps for the fixed and moving images are computed once per resolution level or after a few iterations. Once generated, these feature maps are treated as multi-channel images.

Formally, at each iteration, $N$ spatial coordinates are randomly selected from the fixed image domain $\Omega_F$ (or its corresponding mask). The similarity measure is expressed as follows:\\
\begin{gather}\label{eq:costStatic}
\mathcal{S}(I_F, I_M \circ T_\theta ) = \\ \frac{1}{NC}\sum_{l=1}^L \gamma_l\sum_{i=1}^N \mathcal{D}\big(\phi_{IF_l}(x_i)), \phi_{IM_l}(T_\theta(x_i)\big) \nonumber
\end{gather}
where $\phi_{IF_l} : \Omega_F \rightarrow \mathbb{R^C}$ and $\phi_{IM_l} : \Omega_M \rightarrow \mathbb{R^C}$ represent respectively, from the fixed and moving images, the $C$ features maps extracted at the layer $l$ and $\gamma_l$ represent the weight assigned to that layer.
The gradient expression of the similarity measure (Eq.\ref{eq:gradStatic}) is simplified as there is no need to compute the Jacobian of $\phi$.
\begin{gather}\label{eq:gradStatic}
\frac{\partial \mathcal{S}(I_F, I_M \circ T_\theta)}{\partial \theta} = \\
\frac{2}{NC} \sum_{l = 1}^L \gamma_l \sum_{i = 1}^N \sum_{c = 1}^C 
\left. \frac{\partial \mathcal{D}(f, m)}{\partial m} \right|_{\begin{aligned} 
    f &= \phi^c_{IF_l}(x_i) \\ 
    m &= \phi^c_{IM_l}(T_\theta(x_i)) 
\end{aligned}}\nonumber\\
\times \frac{\partial \phi^c_{I_M}(T_\theta(x_i))}{\partial \theta} \nonumber
\end{gather}

The extracted feature maps are recorded as matrices, however, $\phi_I$ can be interpreted as a continuous function through third-order B-spline interpolation.\\

\subsection{Integration of IMPACT in \textit{Elastix} Framework}

The implementation of the IMPACT within the Elastix framework is highly flexible, allowing for advanced customization. Users can fine-tune the similarity metric parameters to achieve the desired level of precision. 

The deep learning-based feature extraction process is defined by seven parameters: the model path, which specifies where to load the model from; the dimensionality and number of channels of the input patch images fed into the model; the patch size $P$; the processing resolution $R$; a mask for selecting the desired layers within the model to be included in the similarity comparison; and the choice of the distance metric $\mathcal{D}$ for feature comparison. Multiple feature extraction processes from different models can be applied simultaneously. Additionally, users can define a distinct feature extraction setup for each resolution level.

The calculation of $\phi(.)$ (\ref{eq:costJacobian}) on each patch and the computation of Jacobians (\ref{eq:gradJacobian}) are performed using the Torch library \cite{paszke2019pytorch}. The use of TorchScript ensures compatibility with a wide range of pretrained networks, including 2D and 3D DL-based architectures. The system supports parallel computation on the CPU, with tasks divided into batches to accelerate execution across multiple cores. Additionally, GPU support is fully integrated for the computation of $\phi$ and its Jacobian.

To further reduce the computational burden, a mechanism is introduced to randomly sample a subset of features uniformly at each iteration. Instead of comparing all features extracted from the pair of patches, only a randomly selected subset of features is used to compute the similarity metric. In static mode, a Principal Component Analysis (PCA) of all features can also be used to compare a reduced and more compact set of features.

If the number of channels required by the model input differs from that of the image, the missing channels are either duplicated or replaced by the mean value.

When the model's dimensions are smaller than the image, such as when using a pretrained model like SAM2.1 originally designed for natural images in a 3D image registration task, 2D patches are randomly sampled from various planes within the 3D volume. These planes are not restricted to the principal axes (sagittal, coronal, and axial) but are instead randomly defined in the 3D space, allowing for the extraction of slices with diverse orientations. This approach captures the spatial complexity of the image while adapting the data to the 2D model.

In static mode, the extracted feature maps can be obtained via patch inference, where the input image is divided into smaller overlapping patches that are processed independently. This approach is particularly advantageous for handling high resolution 3D images. By dividing each image into patches, the system avoids memory limitations and enables the processing of large datasets. By default, this inference is performed once before each resolution but can be repeated every $k$ iterations.

Please refer to Appendix B for a complete list of metric parameters.

\subsection{Semantic Similarity Metric Using Pretrained Models for Deep Learning-Based Registration}

IMPACT was integrated into the VoxelMorph framework \cite{balakrishnan_voxelmorph_2019} to optimize the learning of a deformable transformation model. VoxelMorph is an unsupervised UNet-based architecture designed to compute a deformation vector field (DVF) for aligning two input images. Its encoder-decoder with skip connections structure allows the encoder to extract multiscale features, while the decoder predicts the DVF. The semantic loss function used in this integration provided two distinct types of features: encoder features, offering general and abstract image representations, and decoder features, optimized for segmentation relevant tasks. Regularization of the DVF is achieved through a diffusion regularizer applied to the spatial gradients, ensuring smooth and diffeomorphic transformations that preserve anatomical topology and continuity. Compared to the Elastix integration, VoxelMorph offers greater flexibility, enabling access to both encoder and decoder features in Jacobian mode while maintaining computational efficiency. This adaptability facilitates efficient exploration of feature representations.
This approach can incorporate embeddings from the same broad set of pretrained models (TotalSegmentator, SAM/MedSAM, STU-Net) without modifying VoxelMorph's main architecture.

\section{Experimental Setup}

This section outlines the experimental protocol used to evaluate the IMPACT metric. Five multimodal registration tasks, involving different imaging modalities (CT, MRI, CBCT) and anatomical regions (thorax, abdomen, and pelvis), are used as benchmarks. Tasks 1, 2, 3, and 5 are addressed using the Elastix framework, while tasks 3 and 4 are addressed using the VoxelMorph framework. The clinical tasks are first presented, followed by the configuration details of both registration frameworks. Finally, the evaluation protocol and the metrics used for quantitative and qualitative assessment are described.

\subsection{Clinical tasks}
\begin{enumerate}  
    \item Thorax CT/CBCT registration (VATSop \cite{Rouze2018}): The task focuses on lung nodule localization during video-assisted thoracoscopic surgery (VATS), a minimally invasive technique used for the surgical treatment of early-stage lung cancer \cite{Rouze2018,alvarez2021hybrid,bou2023}. This dataset is particularly challenging due to inherent CBCT artifacts (scatter, noise, truncation artifacts) \cite{schulze_artefacts_2011} and its limited field of view compared to CT. Additionally, the change in patient positioning from supine to lateral decubitus causes significant lung deformations, including displacements exceeding 40 mm, sliding motions of up to 30 mm between lung lobes and the thoracic wall, and localized volume changes \cite{alvarez2022measurement}. For validation, precise lobe segmentations and approximately forty paired anatomical landmarks (vessel and airway bifurcations) were manually selected per CT/CBCT pair by the VATS surgeon at Rennes University Hospital. This retrospective study received ethical approval (2016-A01353-48 35RC169838), and informed consent was obtained from all patients.
    
    \item Thorax FBCT/CBCT registration (ThoraxCBCT Learn2Reg 2023 MICCAI Challenge \cite{hugo_longitudinal_2017,hering2022learn2reg}): This task focuses on thoracic image registration between pre-therapeutic fan beam CT (FBCT) and interventional CBCT scans \cite{hugo_longitudinal_2017}. The dataset includes three patients, each with longitudinal imaging acquired at different stages of image-guided radiation therapy. For each patient, one FBCT scan is obtained before treatment during maximum inspiration, and two CBCT scans are acquired during maximum expiration. The CBCT images contain substantial artifacts such as scatter, noise, and truncation, which reflect real-world clinical conditions and increase the complexity of the registration task. Evaluation is based on ground truth segmentation masks for lung lobes, tumors, and several organs at risk, including the heart, spinal cord, and esophagus. These annotations are not publicly accessible.
  
    \item Abdomen MR/CT registration (AbdomenMRCT Learn2Reg 2021 MICCAI Challenge \cite{clark_cancer_2013,hering2022learn2reg}): This task involves multimodal registration between abdominal MRI and CT scans. The dataset includes three patients, each with paired volumetric T1-weighted MRI and contrast-enhanced CT scans. Significant modality differences and anatomical variability make the task particularly challenging. Evaluation is based on ground truth segmentation masks for key abdominal structures, including the liver, spleen, kidneys, and pancreas, these annotations are not publicly accessible.
    
    \item Pelvis MR/CT registration (PelvisMRCT \cite{dowling2015automatic}): The task consists of 39 prostate cancer patients, each having both CT and MRI scans. The MRI scans were acquired using 3D T2-weighted SPACE sequences on a 3T Siemens Skyra scanner. Routine planning CT scans were performed using either a GE LightSpeedRT large-bore scanner or a Toshiba Aquilion. Each patient CT scan was coregistered to their whole-pelvis T2 with a robust symmetric rigid registration. For evaluation purposes, manual delineations of the prostate, rectum, bladder, and bones were independently performed on both CT and MRI scans. Ethics approval was obtained from the local area health ethics committee, and informed consent was obtained from all patients.
    
    \item Thorax CT/CT registration (LungCT Learn2Reg 2021 MICCAI Challenge \cite{hering_2020_3835682,hering2022learn2reg}): This task focuses on registering lung CT scans acquired at two respiratory phases—full inspiration and full expiration. The dataset includes eleven patients in total: eight for testing and three for validation. It poses significant challenges due to the large deformations caused by respiratory motion, resulting in pronounced lung displacements and localized volume changes. Additionally, expiration scans have a reduced field of view, further complicating the registration process. Quantitative evaluation is performed based on anatomical landmarks, which are not publicly accessible.
\end{enumerate}

\subsection{Common configuration}
\label{sec:common_config}

The IMPACT metric was integrated as a custom similarity measure and configured with task-specific hyperparameters. Depending on the task, features were extracted from different layers of various pretrained models, with the specific combinations detailed in Appendix A, and compared using different feature distances. In the Elastix framework, two modes of operation were used: Jacobian mode, which enables backpropagation through the feature extractor, and Static mode, which relies on precomputed features and is better suited for extracting deeper semantic representations. The detailed configurations applied to each task are summarized in Table \ref{tab:LossParametersTasks}.
All input images were intensity normalized and resampled into a canonical orientation to ensure compatibility with the pretrained models used for semantic feature extraction, following the specifications detailed in Appendix A.

\begin{table}[h]
    \caption{Hyperparameters used for IMPACT within the Elastix and VoxelMorph frameworks across different tasks. The parameters listed correspond to the official challenge submissions. Additional configurations used in ablation studies are also reported, where "Variable" indicates parameters that were systematically modified for experimental comparison.}
    \centering
    \renewcommand{\arraystretch}{1.2}
\resizebox{0.5\textwidth}{!}{%
    \begin{tabular}{|c|c|ccccc|}
    \hline
    Framework & Task & Model & PatchSize & Layers & Distance & Mode\\
    \hline
    \multirow{6}{*}{Elastix}&1 & M258 & 11*11*11 & 2 &L2 &Jacobian\\
    &1(Ablation) & Variable & Variable & Variable & Variable &Jacobian\\
    &2 & M730-731 & 11*11*11 & 2 & L1 & Jacobian\\
    &3 & M730 & - & 8 &  L1 &Static\\
    &5 & M258 & 11*11*11& 2 & L2 &Jacobian\\
    &5(Ablation) & Variable & Variable & Variable & Variable &Jacobian\\
     \hline
    \multirow{2}{*}{Voxelmorph}& 3 & M730 & - & 7 & L1 & Jacobian\\
    &4 & M730 & - & Variable & L1 & Jacobian\\
    \hline
    \end{tabular}
    }
    \label{tab:LossParametersTasks}
\end{table}

\subsection{Elastix setup}
\label{sec:elastix_proto}

All Elastix-based experiments shared a common configuration for deformable registration. A 3D B-spline transformation model was optimized using the ASGD optimizer \cite{Klein2008}. The similarity measure was computed at each iteration over a set of 2000 points randomly sampled in physical coordinates. The use of 2000 sampled points is sufficient to provide a reliable estimate of image similarity and to ensure stable convergence. This was combined with a four-level multi-resolution strategy, where image resolution was progressively refined through resampling at spacings of 6 mm, 3 mm, 1.5 mm, and 1 mm. Simultaneously, the B-spline control point spacing was halved at each resolution level, reaching a final spacing of 8 mm at the finest resolution. This configuration was chosen as a robust default across tasks and is further justified by the analysis in Appendix C.

A weakly supervised strategy was employed in tasks 1 and 5 by restricting the similarity computation to the lung region of the fixed image. Task 3 included an affine initialization step prior to non-rigid registration, using the same IMPACT configuration but with an affine transformation model. All experiments were conducted on a workstation equipped with an Intel Core i7-12700K CPU, 32 GB of RAM, and an NVIDIA RTX 3080 GPU.

\subsection{VoxelMorph setup}
\label{sec:vxm_proto}

A vanilla VoxelMorph model was implemented in PyTorch and trained using the AdamW optimizer with a learning rate of 0.0002, processing one image pair per iteration. The model was trained for 200 epochs on an NVIDIA RTX A6000 GPU. VoxelMorph was used as an optimization tool rather than a generalizable model, training was therefore performed directly on the image pairs of each task, without the need for a separate training set. As in the Elastix-based setup, task 3 included an initial affine alignment step using \textit{Elastix}.

\subsection{Evaluation Protocol}
The evaluation protocol was designed to thoroughly assess the registration performance of the proposed methods across a variety of tasks. Table \ref{tab:task_method_results} provides a quick overview of which figures and metrics correspond to which task and method.

For task 1, the efficiency of IMPACT was demonstrated by comparing its performance in terms of Target Registration Error (TRE), Dice Similarity Coefficient (DSC) and the 95th percentile Hausdorff distance (HD95) against various conventional similarity measures, including MSE, NCC, NMI and MIND. For a fair comparison, an extraction model for the MIND features has been developed, enabling it to be used in algorithmic registration with Elastix. Four MIND configurations were tested with a radius of 1 and 2, as well as a dilation of 1 and 2. Only the best-performing configuration is presented. Moreover, the convergence speed based on the number of iterations was evaluated, exploring several values: [50, 100, 150, 200, 500, 1000, 1500, 2000]. Tasks 2 and 3 were evaluated primarily using challenge-specific metrics, including the DSC, HD95, and TRE, which was used exclusively for Task 2. For Task 4, registration accuracy was assessed using only the DSC.

For task 4, the model was trained on the full pelvic MR/CT dataset. Different semantic feature configurations were evaluated, including encoder and decoder layers from TotalSegmentator and the features of the second layer of MedSAM3D. As baselines, MI and MIND descriptors were also integrated into the same VoxelMorph framework to enable direct comparisons with conventional similarity metrics.

Qualitative evaluations were conducted across tasks 1 and 4 to complement quantitative results. This involved visual inspections to assess alignment quality, contour accuracy, and the presence of noticeable artifacts, ensuring the robustness and interpretability of the results.

Additionally, an ablation study was conducted on tasks 1 and 5 to assess the impact of different hyperparameter choices on the performance of IMPACT, evaluated using TRE.

Finally, a computational complexity analysis was performed to evaluate execution time. This holistic evaluation framework ensures a well-rounded assessment of both algorithmic and learning-based approaches.

\begin{table}[ht]
\caption{Method result per task mapping with relevant figures}
\begin{tabular}{@{}>{\raggedright\arraybackslash}p{2cm} >{\raggedright\arraybackslash}p{2cm} >{\raggedright\arraybackslash}p{2cm} >{\raggedright\arraybackslash}p{1.5cm}@{}}
\toprule
\textbf{Task} & \textbf{Methodology (Section)} & \textbf{Evaluation Metric} & \textbf{Figure(s)} \\ \midrule
Task 1: VATSop & Elastix (Section \ref{sec:elastix_proto}) & Qualitative, TRE, DSC, HD95 & Fig. \ref{fig:qualitative_task1},\ref{fig:boxplot_TRE_task1}, Table \ref{tab:results_task1} and \ref{tab:ablation_task1}\\ 
Task 2: ThoraxCBCT & Elastix (Section \ref{sec:elastix_proto}) & DSC, TRE & Table \ref{tab:results_task2}\\ 
Task 3: AbdomenMRCT & Elastix \& VoxelMorph (Section \ref{sec:elastix_proto}-\ref{sec:vxm_proto}) & Qualitative, DSC & Fig. \ref{fig:qualitative_task3} and Table \ref{tab:results_task3}  \\ 
Task 4: PelvisMRCT & VoxelMorph (Section \ref{sec:vxm_proto}) & DSC & Fig. \ref{fig:boxplot_dice_task4} \\ 
Task 5: LungCT & Elastix (Section \ref{sec:elastix_proto}) & TRE & Table \ref{tab:results_task5} and \ref{tab:ablation_task5} \\ 

\bottomrule
\end{tabular}
\label{tab:task_method_results}
\end{table}

\section{Results}

\subsection{Qualitative results}
\begin{figure}[h!]
    \centering
    \includegraphics[width=0.49\textwidth]{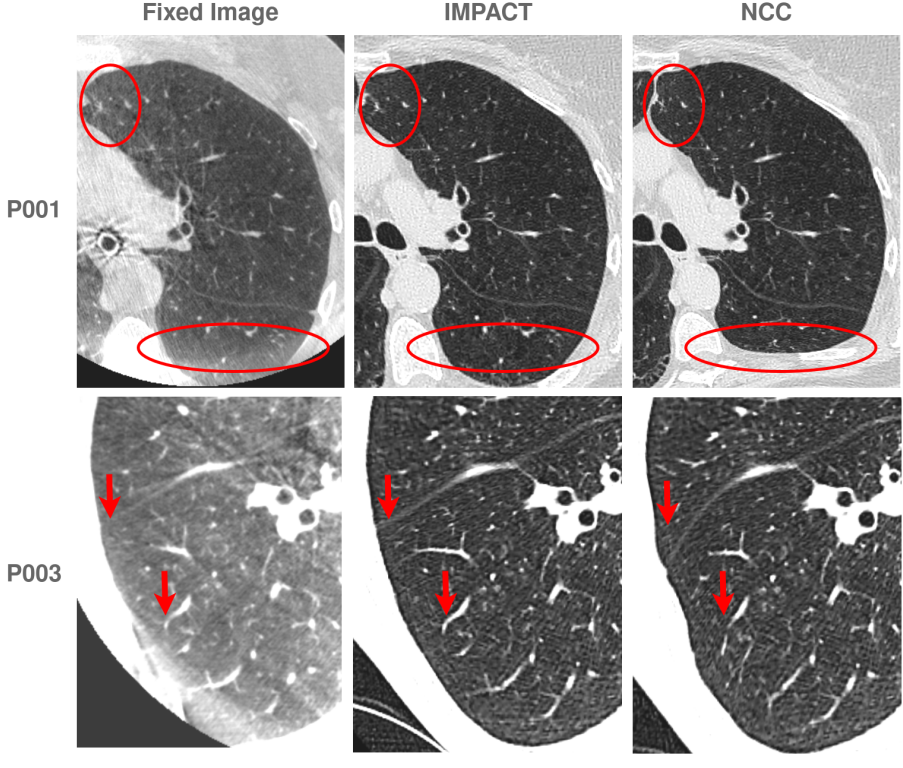}
    \caption{Qualitative comparison of the image registration results. The reference image (first column) is compared with the registered moving image obtained after 2000 iterations (presented in the second and third columns). The registered image obtained using IMPACT is presented in the second column and the one obtained via NCC in the third. Each row presents an example from two distinct patients.}
    \label{fig:qualitative_task1}
\end{figure}

Qualitative registration outcomes for Task 1 are presented in Fig. \ref{fig:qualitative_task1}. Two cropped regions are shown (each row corresponds to a different patient): the first column displays the fixed image (CBCT), while the second and third columns show the registered moving image (CT) using IMPACT and NCC, respectively, after 2000 iterations. Notably, when guided by our metric, the lung contours and local structures (circled or indicated with arrows) exhibit visibly improved alignment, even in the presence of CBCT artifacts.

\begin{figure*}[h!]
    \centering
    \includegraphics[width=0.7\textwidth]{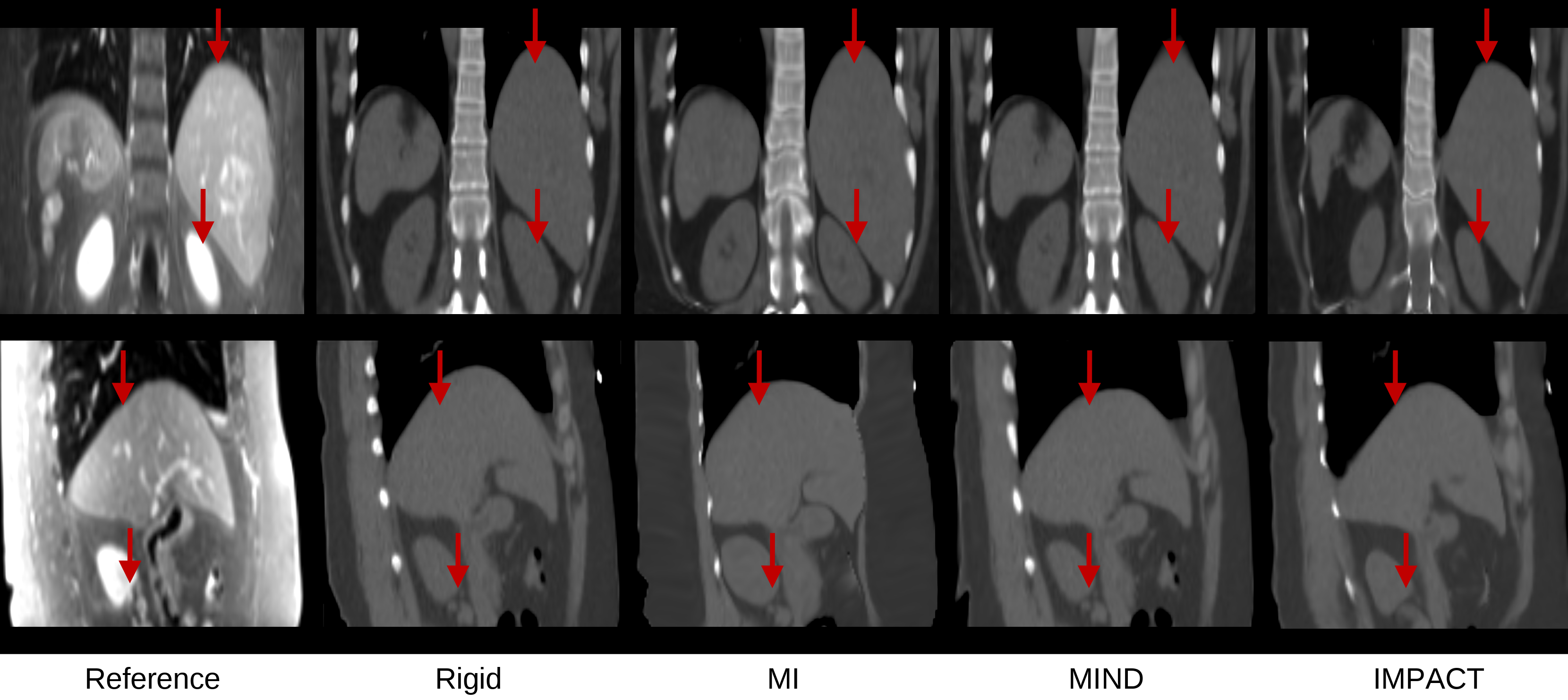}
    \caption{Qualitative results obtained using VoxelMorph on Task 3 with different loss functions. The first column presents the reference image, followed by results from rigid alignment (Rigid) and the registered CT images estimated using three loss functions: MI, MIND, and the proposed IMPACT loss. Each row illustrates a different example, highlighting the influence of the loss function on registration quality.}
    \label{fig:qualitative_task3}
\end{figure*}

Fig. \ref{fig:qualitative_task3} shows abdomen MR/CT registration results (Task 3) using VoxelMorph with three different similarity metrics: MI, MIND, and our proposed IMPACT. The first column is the reference (fixed) MRI, followed by a rigidly aligned CT, and then the warped CT with MI, MIND, and IMPACT. As the figure shows, MI and MIND-based registrations sometimes fail to align organs with high precision, whereas IMPACT captures more anatomically coherent correspondence.

\subsection{Quantitative evaluation and convergence analysis}
\begin{figure}[h!]
    \centering
    \includegraphics[width=0.5\textwidth]{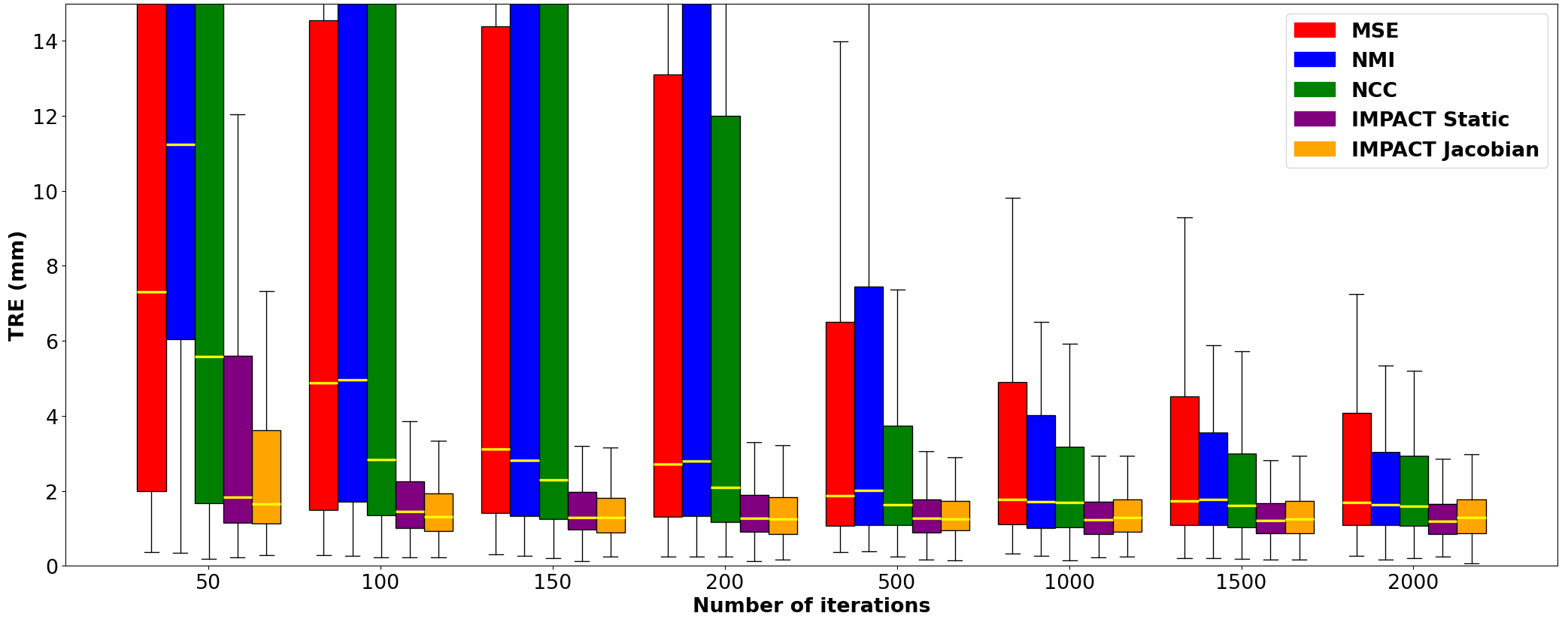}
    \caption{Registration performance measured in terms of TRE. The box plots represent the distribution of errors across varying numbers of iterations, comparing different cost functions: MSE, NMI, NCC, and IMPACT in both Jacobian and Static modes.}
    \label{fig:boxplot_TRE_task1}
\end{figure}

Figure \ref{fig:boxplot_TRE_task1} illustrates an experiment designed to evaluate the convergence behavior and accuracy of different similarity metrics during registration. We compare MSE, NMI, NCC, and IMPACT (in both Jacobian and Static modes) on Task 1. The plot shows the TRE distribution as a function of the number of iterations. IMPACT not only converges faster but also achieves the lowest median TRE with a notably tighter interquartile range, indicating more consistent performance across patients.

Table \ref{tab:results_task1} summarizes the registration performance on Task 1 after 2000 iterations, comparing different similarity measures using TRE, DSC, and HD95. IMPACT in Jacobian mode achieves the best overall performance. The Static variant performs similarly, with slightly reduced boundary accuracy. Notably, IMPACT outperforms the strongest baseline, MIND, across all evaluation metrics. Traditional metrics MSE, NMI, and NCC perform substantially worse. The rigid case, used as initialization, reflects the absence of deformation and highlights the large displacements to be estimated by the deformable models.

\begin{table}[h!]
\caption{Evaluation of different similarity metrics on Task 1 after 2000 iterations, using TRE, DSC, and HD95.}
\centering
\footnotesize
\renewcommand{\arraystretch}{1.2}
\resizebox{0.49\textwidth}{!}{%
\begin{tabular}{|l|c|c|c|c|c|}
\hline
Method                  & TRE (25\%) & TRE (50\%) & TRE (75\%) & DSC  & HD95 \\ \hline
Rigid (Before Registration) & 9.44           & 13.59           & 17.58           & 0.79    & 14.75         \\ \hline
MSE                     & 1.09           & 1.68  & 4.08  & 0.91 & 9.91      \\ \hline
NMI           & 1.09 & 1.62  & 3.03  & 0.85 & 14.54     \\ \hline
NCC           & 1.06 & 1.58  & 2.92  & 0.88 & 12.01     \\ \hline
MIND (R1D2)          & 0.93 & 1.25 & 1.85 & 0.96 & 3.70 \\ \hline
\textbf{IMPACT Jacobian}         & \textbf{0.85} & \textbf{1.2}  & \textbf{1.71}  & \textbf{0.97} & \textbf{3.19}      \\ \hline
IMPACT Static & 0.85 & 1.22  & 1.71  & 0.95 & 6.06      \\ \hline
\end{tabular}%
}

\label{tab:results_task1}
\end{table}

Elastix with IMPACT was used in the Learn2Reg ThoraxCBCT challenge, which involves highly artifact-prone CBCT images acquired across different respiratory phases. As shown in Table \ref{tab:results_task2}, our method (team “BreizhReg”) ranked 6th among all participants.

\begin{table*}[ht]
    \caption{Ranking results of the 21 participants in \href{https://learn2reg.grand-challenge.org/evaluation/thoraxcbct/leaderboard/}{Learn2Reg ThoraxCBCT Challenge}. The table presents results based on DSC. The non-rigid registration based on Elastix with Impact ranked 6th.}
    \centering
    \renewcommand{\arraystretch}{1}
    \setlength{\tabcolsep}{6pt}
    \resizebox{0.6\textwidth}{!}{%
    \begin{tabular}{@{}clcc@{}}
        \toprule
        \textbf{\#} & \textbf{User (Team)} & \textbf{DSC} \\ \midrule
        \textbf{1st} & \textcolor{black}{\textbf{mysterious\_man}} & \textbf{0.732 $\pm$ 0.0026} \\ 
        \textbf{2nd} & \textcolor{black}{anonymous2024} & 0.689 $\pm$ 0.0053  \\
        \textbf{3rd} & \textcolor{black}{SSKJLBW} & 0.669 $\pm$ 0.0021  \\
        \textbf{6th} & \textcolor{cyan}{Our team (BreizhReg)} \textit{(Elastix)} & \textcolor{cyan}{0.652 $\pm$ 0.0032}  \\
        \textbf{8th} & \textcolor{black}{Challenge Organizers}\textit{(deedsBCV)}  & 0.648 $\pm$ 0.0030   \\

        \textbf{12th} & \textcolor{black}{Challenge Organizers}\textit{(ConvexAdam 30 label)}  & 0.586 $\pm$ 0.0027 \\
        \textbf{15th} & \textcolor{black}{Challenge Organizers}\textit{(NiftyReg)}  & 0.568 $\pm$ 0.0036  \\
        \textbf{17th} & \textcolor{black}{Challenge Organizers}\textit{(Voxelmorph++)}  & 0.503 $\pm$ 0.113  \\
        \textbf{21th} & \textcolor{black}{Challenge Organizers}\textit{(Initial Displacements)}  & 0.313 $\pm$ 0.093  \\
        \bottomrule
    \end{tabular}%
    }
    \label{tab:results_task2}
\end{table*}

The Abdomen MRCT task of the Learn2Reg 2021 MICCAI Challenge, characterized by strong intensity and contrast differences between T1-weighted MRI and CT, was used to evaluate both Elastix and VoxelMorph frameworks with IMPACT. As shown in Table \ref{tab:results_task3}, Elastix ranked 7th overall, while VoxelMorph achieved 2nd place among 26 participants (average DSC: 0.8973, HD95: ~3.32 mm). Although both use the same similarity metric, they rely on different deformation models. VoxelMorph predicts DVFs through a learning-based approach trained on the dataset. In contrast, Elastix uses a B-spline transformation with strong regularization, estimated independently for each image pair.

\begin{table*}[ht]
\caption{Ranking results of the 26 participants in \href{https://learn2reg.grand-challenge.org/evaluation/test/leaderboard/}{Task 1 (Abdomen MRCT) of the Learn2Reg 2021 Challenge}. The table presents results based on DSC, and HD95 metrics where higher DSC values and lower HD95 values indicate better performance in medical image registration. The non-rigid registration based on Elastix with IMPACT ranked 7th and non-rigid registration based on VoxelMorph with IMPACT ranked 2nd.}
    \centering
    \renewcommand{\arraystretch}{1}
    \setlength{\tabcolsep}{6pt}
    \resizebox{0.7\textwidth}{!}{%
    \begin{tabular}{@{}clccc@{}}
        \toprule
        \textbf{\#} & \textbf{User (Team)} & \textbf{DSC} & \textbf{HD95} \\ \midrule
        \textbf{1st} & \textcolor{black}{\textbf{cwmokab (Orange)}} & \textbf{0.9148 $\pm$ 0.0018} & \textbf{2.7269} \\ 
        \textbf{2nd} & \textcolor{cyan}{Our team (BreizhReg)} \textit{(Voxelmorph)} & \textcolor{cyan}{0.8973} $\pm$ \textcolor{cyan}{0.0152} & \textcolor{cyan}{3.3169} \\ 
        \textbf{3rd} & \textcolor{black}{honkamj} & 0.8956 $\pm$ 0.0137 & 2.8451 \\ 
        \textbf{6th} & \textcolor{black}{Challenge Organizers} \textit{(corrField)} & 0.8757 $\pm$ 0.0168 &  4.4036 \\ 
        \textbf{7th} & \textcolor{cyan}{Our team (BreizhReg)} \textit{(Elastix)} & \textcolor{cyan}{0.8748} $\pm$ \textcolor{cyan}{ 0.0066} &  \textcolor{cyan}{3.1051} \\ 
        \textbf{13th} & \textcolor{black}{Challenge Organizers} \textit{(AdamReg Square MIND)} & 0.8040 $\pm$ 0.0253 & 7.0\\ 
        \textbf{19th} & \textcolor{black}{Challenge Organizers} \textit{(NiftyReg MIND)} & 0.5395 $\pm$ 0.1593 &  15.0666 \\ 
        \textbf{24th} & \textcolor{black}{Challenge Organizers} \textit{(Initial Displacements)} & 0.3096 $\pm$ 0.1556 &  22.8887 \\ 
        \bottomrule
    \end{tabular}%
    }
    
    \label{tab:results_task3}
\end{table*}

\begin{figure}[h!]
     \includegraphics[width=0.5\textwidth]{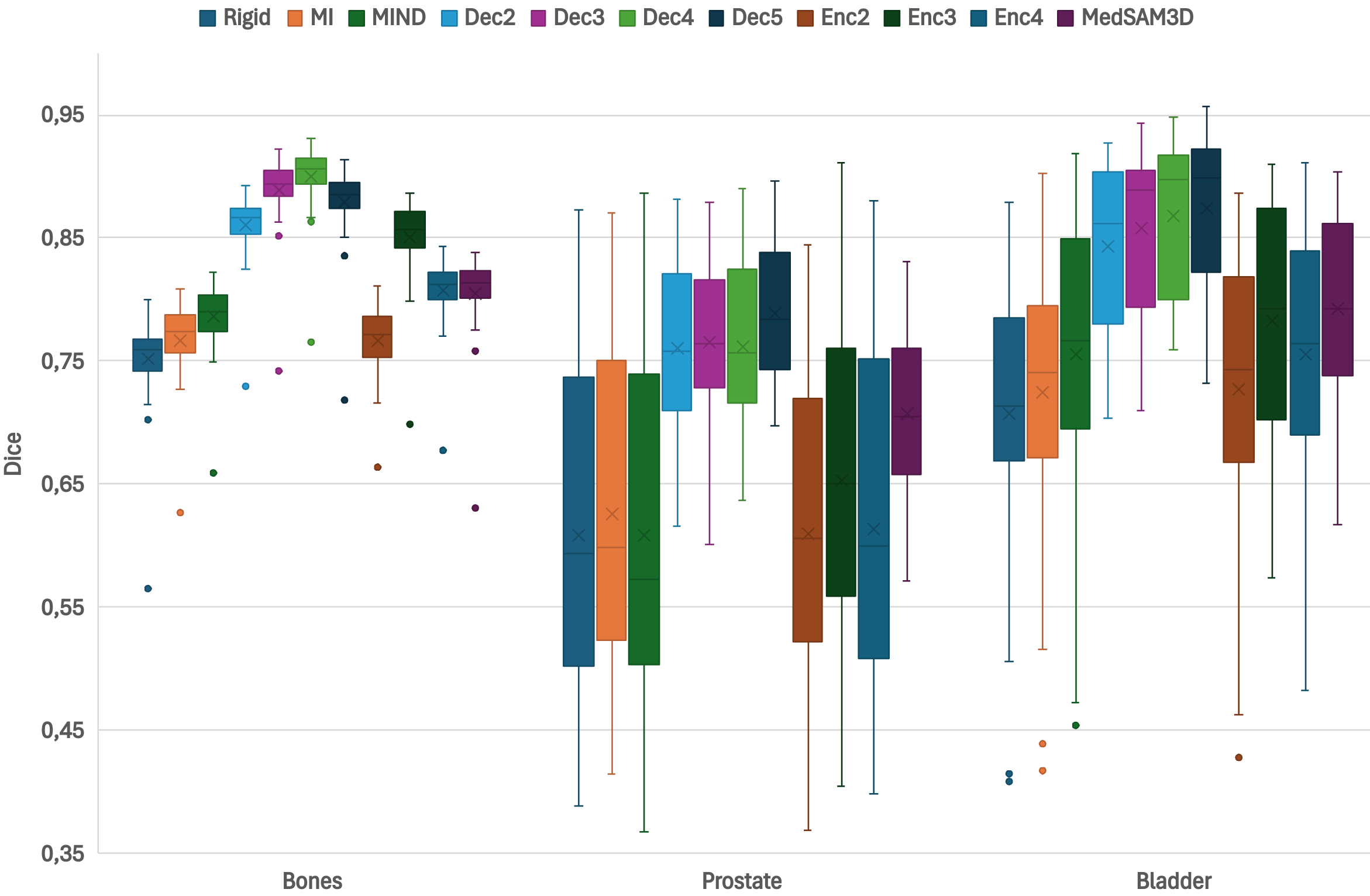}
    \caption{Boxplot of DSC per organ for rigid registration and non-rigid registration using VoxelMorph with different loss functions, including MI, MIND, and IMPACT. The figure also presents DSC values for IMPACT when applied with feature representations from layers 2, 3, and 4 of the decoder (Dec) and encoder (Enc) of the TotalSegmentator model M730, and the second layer of the MedSAM3D network.}
    \label{fig:boxplot_dice_task4}
\end{figure}

Fig. \ref{fig:boxplot_dice_task4} compares TotalSegmentator encoder and decoder layers, MedSAM3D features, and traditional similarity metrics (MI, MIND) within the VoxelMorph registration framework. Overall, the encoder layers from TotalSegmentator yield results roughly equivalent to MI and MIND, except for layer 3, which surpasses these metrics. Minor variations do appear among different decoder depths, features based on decoder layers outperform all other metrics. Meanwhile, MedSAM3D features produce Dice scores better than MI and MIND but still lower than those achieved by TotalSegmentator decoder layers. 

\begin{table*}[ht]
\caption{Ranking results of the 32 participants in \href{https://learn2reg.grand-challenge.org/evaluation/task-1-validation/leaderboard/}{Task 2 (LungCT) of the Learn2Reg 2021 Challenge}.
The table presents results based on TRE. Elastix with IMPACT ranked 3rd in this leaderboard.}
    \centering
    \renewcommand{\arraystretch}{1}
    \setlength{\tabcolsep}{6pt}
    \resizebox{0.6\textwidth}{!}{%
    \begin{tabular}{@{}clccc@{}}
        \toprule
        \textbf{\#} & \textbf{User (Team)} & \textbf{TRE} \\ \midrule
        \textbf{1st} & \textcolor{black}{Challenge Organizers} \textit{(Fraunhofer MEVIS)}& \textbf{1.8356 $\pm$ 0.5685}\\ 
        \textbf{2nd} & \textcolor{black}{Challenge Organizers} \textit{(corrField)} & 1.8497 $\pm$ 0.7033\\ 
        \textbf{3rd} & \textcolor{cyan}{Our team (BreizhReg)} \textit{(Elastix)} & \textcolor{cyan}{1.8986} $\pm$ \textcolor{cyan}{0.7207} \\
        \textbf{11th} & \textcolor{black}{Challenge Organizers} \textit{SLIC-Reg++)}& 2.6359 $\pm$ 1.0163\\ 
        \textbf{21th} & \textcolor{black}{Challenge Organizers} \textit{(Initial Displacements)}& 14.6407 $\pm$ 6.0765	\\ 
        \bottomrule
    \end{tabular}%
    }
    \label{tab:results_task5}
\end{table*}
We further evaluated Elastix with IMPACT on the LungCT task of the Learn2Reg 2021 Challenge, which focuses on aligning lung CT scans using anatomical landmarks as evaluation targets. As reported in Table \ref{tab:results_task5}, our method (team “BreizhReg”) achieved a 3rd place ranking out of 32 participants, with a mean TRE of 1.90 mm.

\subsection{Computational Complexity Analysis}

The computational cost of the method depends linearly on the number of iterations $N$, the number of sampled points per iteration $S$, and the cost $f(.)$ of computing the similarity on a single sample. The overall complexity can be expressed as:
\begin{equation}
    \mathcal{O}(N.S.f(.))
\end{equation}

An estimate of the total runtime can be directly inferred from this relation for a given configuration. Table \ref{tab:computational_cost} provides empirical runtimes for a representative setting with $N=500$ and $S=2000$, covering both standard similarity metrics and IMPACT using different pretrained models, including MIND, TotalSegmentator (TS), and SAM2.1, in both static and Jacobian modes for the first and second feature layer.

\begin{table}[h] 
\caption{Empirical computational cost for different similarity metrics, evaluated with \(N=500\) iterations and \(S=2000\) samples. For each configuration, the runtime in seconds is reported for the first and second feature layers.}
\centering
\footnotesize
\renewcommand{\arraystretch}{1.2}
\setlength{\tabcolsep}{4pt}
\resizebox{0.43\textwidth}{!}{%
\begin{tabular}{|c|c|c|c|} 
\hline
\multicolumn{4}{|c|}{\textbf{Intensity-based Metrics}} \\
\hline
\multicolumn{2}{|c|}{MSE (s)} & \multicolumn{1}{c|}{NCC (s)} & \multicolumn{1}{c|}{NMI (s)} \\ \hline
\multicolumn{2}{|c|}{15.10} & \multicolumn{1}{c|}{16.45} & \multicolumn{1}{c|}{114.62} \\ \hline
\multicolumn{4}{|c|}{\textbf{Per-Model Evaluation}} \\
\hline
Model & Mode & First Layer (s) & Second Layer (s)\\ 
\hline
\multirow{2}{*}{MIND} & Jacobian & 230.98 & -- \\
 & Static & 105.29 & -- \\
\hline
\multirow{2}{*}{TS} & Jacobian & 131.55 & 586.72 \\
 &Static & 160.83 & 170.56 \\
\hline
SAM2.1 & Jacobian & 393.58 & 408.87 \\
\hline
\end{tabular}
}
\label{tab:computational_cost}
\end{table}

\section{Ablation studies}
To analyze the impact of different hyperparameter choices on the performance of IMPACT, an ablation study was conducted on tasks 1 and 5. This experiment aims to evaluate three key parameters in terms of TRE : (1) the pretrained model used for feature extraction, (2) the distance measure between extracted features, and (3) the feature extraction level within the model.

\begin{table*}[h!]
\caption{Quantitative comparison of pretrained models, feature distance metrics, and feature extraction levels used within IMPACT on Task 1, reporting the mean and standard deviation of the TRE. Intensity-based metrics are also reported for reference.}
\centering
\footnotesize
\renewcommand{\arraystretch}{1.2}
\setlength{\tabcolsep}{4pt} 
\resizebox{\textwidth}{!}{
\begin{tabular}{|c|c|c|c|c|c|c|c|c|}
\hline
\multicolumn{9}{|c|}{\textbf{Intensity-based metrics}} \\ \hline
\multicolumn{3}{|c|}{MSE} & \multicolumn{3}{c|}{NCC} & \multicolumn{3}{c|}{NMI} \\ \hline
\multicolumn{3}{|c|}{1.86 $\pm$ 4.97} & \multicolumn{3}{c|}{1.79 $\pm$ 3.68} & \multicolumn{3}{c|}{1.78 $\pm$ 4.61} \\ \hline
\multicolumn{9}{|c|}{\textbf{Per-Model Evaluation}} \\ \hline
\multirow{2}{*}{\textbf{Model}} & \multicolumn{4}{c|}{First Layer} & \multicolumn{4}{c|}{Second Layers} \\ \cline{2-9}
 & L1 & L2 & NCC & Cosine & L1 & L2 & NCC & Cosine \\ \hline
MIND (R1D2) & 1.28 $\pm$ 0.97 & \textbf{1.27 $\pm$ 0.89} & 1.26 $\pm$ 0.95& 1.32 $\pm$ 0.97  & & & &\\
\hline
M258 & 1.42 $\pm$ 1.53 & 1.58 $\pm$ 5.08 & 1.29 $\pm$ 0.86 & 1.7 $\pm$ 8.56 &1.23 $\pm$ 0.87 & \textcolor{cyan}{\textbf{1.2 $\pm$ 0.82}} & 1.25 $\pm$ 0.86 & 1.26 $\pm$ 0.89\\
M291 & 1.23 $\pm$ 0.86 & 1.25 $\pm$ 0.87 & 1.3 $\pm$ 0.89 & 1.24 $\pm$ 0.99 & 1.24 $\pm$ 0.94 & \textbf{1.22 $\pm$ 0.92} & 1.32 $\pm$ 1.02 & 1.23 $\pm$ 0.88\\
M730 & 1.23 $\pm$ 0.93 & \textbf{1.2 $\pm$ 0.84} & 1.22 $\pm$ 0.86 & 1.27 $\pm$ 0.91 & 1.22 $\pm$ 0.84 & 1.2 $\pm$ 0.84 & 1.3 $\pm$ 1.02 & 1.22 $\pm$ 0.91\\
\hline
ConvNeXt & \textbf{1.25 $\pm$ 0.86} & 1.44 $\pm$ 1.25 & 1.51 $\pm$ 2.51 & 1.35 $\pm$ 1.14 & 1.31 $\pm$ 1.01 & 2.19 $\pm$ 4.67 & 2.29 $\pm$ 6.02 & 2.44 $\pm$ 4.86\\ \hline
SAM2.1   &  1.23 $\pm$ 0.97& 1.34 $\pm$ 1.17 & 1.48 $\pm$ 1.74 & 1.39 $\pm$ 1.48 & \textbf{1.22 $\pm$ 0.82} & 1.27 $\pm$ 0.84 & 1.25 $\pm$ 0.89 & 1.26 $\pm$ 0.84\\
MedSAM2  & 1.27 $\pm$ 0.97 & 1.34 $\pm$ 1.15 & 1.49 $\pm$ 1.89 & 1.56 $\pm$ 1.97 & 1.25 $\pm$ 0.83 & \textbf{1.23 $\pm$ 0.84} & 1.23 $\pm$ 0.84 & 1.23 $\pm$ 0.83\\ \hline
\end{tabular}%
.
}
\label{tab:ablation_task1}
\end{table*}

\begin{table*}[h!]
\caption{Quantitative comparison of pretrained models, feature distance metrics, and feature extraction levels used within IMPACT on Task 5, reporting the mean and standard deviation of the TRE. Intensity-based metrics are also reported for reference.}
\centering
\footnotesize
\renewcommand{\arraystretch}{1.2}
\setlength{\tabcolsep}{4pt} 
\resizebox{\textwidth}{!}{%
\begin{tabular}{|c|c|c|c|c|c|c|c|c|}
\hline
\multicolumn{9}{|c|}{\textbf{Intensity-based metrics}} \\ \hline
\multicolumn{3}{|c|}{MSE} & \multicolumn{3}{c|}{NCC} & \multicolumn{3}{c|}{NMI} \\ \hline
\multicolumn{3}{|c|}{9.63 $\pm$ 10.37} & \multicolumn{3}{c|}{5.15 $\pm$ 8.2} & \multicolumn{3}{c|}{2.91 $\pm$ 3.77} \\ \hline
\multicolumn{9}{|c|}{\textbf{Per-Model Evaluation}} \\ \hline
\multirow{2}{*}{\textbf{Model}} & \multicolumn{4}{c|}{First Layer} & \multicolumn{4}{c|}{Second Layer} \\ \cline{2-9}
 & L1 & L2 & NCC & Cosine & L1 & L2 & NCC & Cosine \\ \hline

MIND (R1D2) & \textbf{1.75 $\pm$ 3.18} & 1.88 $\pm$ 3.73 & 1.9 $\pm$ 3.82 & 25.63 $\pm$ 32.67 & & & & \\ 
 \hline
M258 & 1.95 $\pm$ 2.91 & 1.94 $\pm$ 2.82 &2.11 $\pm$ 2.84 & 2.35 $\pm$ 3.35 & 1.36 $\pm$ 1.87 & \textcolor{cyan}{\textbf{1.32 $\pm$ 1.83}} & 1.42 $\pm$ 1.97 & 1.34 $\pm$ 1.92\\
M291  & 2.07 $\pm$ 2.96 & 2.36 $\pm$ 3.85 &2.9 $\pm$ 3.67 & 3.31 $\pm$ 3.86 & 1.66 $\pm$ 2.6 & \textbf{1.62 $\pm$ 2.68} & 1.69 $\pm$ 2.44 & 3.41 $\pm$ 5.69\\
M730  & 1.5 $\pm$ 2.25 & 1.4 $\pm$ 2.07 &1.73 $\pm$ 2.44 & 1.81 $\pm$ 2.6 & \textbf{1.43 $\pm$ 2.01} & 1.47 $\pm$ 2.32 & 1.56 $\pm$ 2.41 & 1.69 $\pm$ 3.18\\
\hline
ConvNeXt & 9.12 $\pm$ 5.7 & 8.07 $\pm$ 7.78 & 7.45 $\pm$ 6.78 & 9.12 $\pm$ 5.7 & \textbf{7.14 $\pm$ 5.78} & 7.71 $\pm$ 6.25 & 7.42 $\pm$ 5.84 & 8.19 $\pm$ 5.66 \\ \hline
SAM2.1   & 2.52 $\pm$ 4.45 & 2.59 $\pm$ 4.95 & 2.5 $\pm$ 4.56 & 6.19 $\pm$ 9.28 & \textbf{1.48 $\pm$ 2.37} & 1.56 $\pm$ 2.58 &  1.49 $\pm$ 2.29 & 1.74 $\pm$ 2.85 \\
MedSAM2  & 2.49 $\pm$ 4.38 & 2.62 $\pm$ 5.18 & 2.53 $\pm$ 4.52 & 6.16 $\pm$ 9.0 & \textbf{1.48 $\pm$ 2.36} & 1.57 $\pm$ 2.59 & 1.5 $\pm$ 2.3 & 1.72 $\pm$ 2.83 \\\hline
\end{tabular}%
}
\label{tab:ablation_task5}
\end{table*}

\subsection{Impact of the Pretrained Model}
Several pretrained models were evaluated to assess their influence on the quality and relevance of extracted features for the registration tasks. The models included ConvNeXt, SAM2.1, MedSAM2, and the TotalSegmentator variants M258, M291, and M730 (see Appendix A for details on training data and target tasks).

Tables \ref{tab:ablation_task1} and \ref{tab:ablation_task5} indicate that M258 consistently achieves the lowest TRE across Tasks 1 and 5, particularly when using second-layer features combined with the L2 distance, yielding TREs of $1.20 \pm 0.82$ and $1.32 \pm 1.83$, respectively. Other segmentation-based models, including M291, M730, SAM2.1, and MedSAM2, also deliver strong performance, though with slightly higher variability. In contrast, the classification-based model ConvNeXt consistently shows the weakest results, with TREs reaching $1.25 \pm 0.86$ on Task 1 and $7.14 \pm 5.78$ on Task 5 under its best configuration. Finally, the handcrafted descriptor MIND, which we reimplemented as a pretrained model within the IMPACT optimization scheme, shows competitive performance in Task 1, especially with L2 distance. However, it is consistently outperformed by learned features on Task 5, even in its best configuration (R1D2).

\subsection{Impact of the Distance Measure}

We analyzed how different distance metrics affect registration performance by evaluating their ability to emphasize orientation, magnitude, or a combination of both within the feature space. The compared metrics included cosine similarity, NCC, and L1/L2 distances.

Cosine similarity captures only the angular relationship between feature vectors and is invariant to scale. While this may be beneficial in tasks where direction alone is informative, it disregards magnitude differences that often encode essential anatomical variations. As a result, cosine consistently underperformed across tasks, particularly in Task 5 (Table~\ref{tab:ablation_task5}).

NCC incorporates both orientation and relative magnitude and generally performs well when features are linearly related. However, in our experiments, it showed inconsistent improvements and higher variability across models and tasks, limiting its robustness.

In contrast, L1 and L2 distances, which measure absolute differences in feature intensity, systematically achieved the best results across all configurations (Tables \ref{tab:ablation_task1} and \ref{tab:ablation_task5}). This was especially pronounced when using deeper semantic features.

\subsection{Impact of the Feature Extraction Level}

\begin{figure*}[h!]
    \centering
    \includegraphics[width=0.7\textwidth]{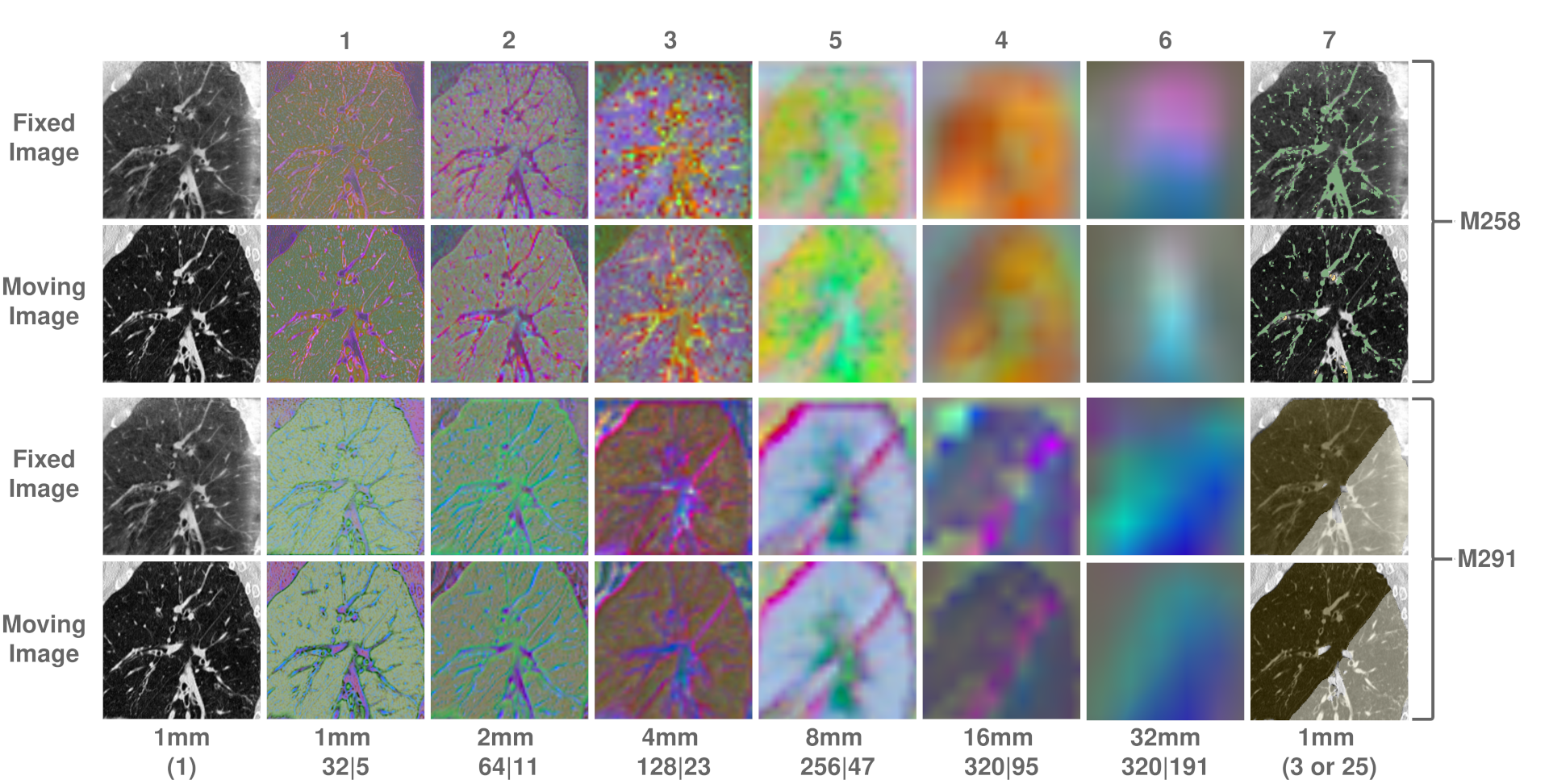}
    \caption{Visualization of the aligned fixed and moving images (first column), feature maps extracted at each encoder stage of the nnU-Net encoder (columns 1–6), and the Softmax output (column 7) for the pretrained models M258 and M291. The feature maps are displayed as RGB images, where the three components are obtained using PCA, retaining the top three principal components for visualization. The resolution (in mm), the number of features, and the receptive field size are indicated below each column in the format resolution$\mid$receptive field (as $32\mid5$). This representation highlights the hierarchical feature extraction process, with increasing spatial context and abstraction at successive encoder stages.}
    \label{fig:featuresMap}
\end{figure*}

Figure \ref{fig:featuresMap} illustrates the hierarchical nature of feature extraction within the nnU-Net encoder. The feature maps, visualized as RGB images using PCA, show the progressive abstraction from fine-grained spatial details in early layers to more semantic representations in deeper layers. To evaluate the impact of this hierarchy on registration performance, features from the first and second encoder layers were tested independently. As reported in Tables \ref{tab:ablation_task1} and \ref{tab:ablation_task5}, the extraction level significantly influences TRE. In most cases, features from the second layer lead to lower errors and more stable results compared to those from the first layer.

\section{Discussion}

This work introduced IMPACT, a semantic similarity metric for multimodal medical image registration that leverages semantic features extracted from large-scale pretrained segmentation models as a generic, modality-agnostic feature space. The metric was integrated into both algorithmic and deep learning–based registration frameworks, and evaluated across five diverse tasks involving different anatomical regions (thorax, abdomen, pelvis) and imaging modalities (CT, CBCT, MRI).

Experimental results demonstrate that semantic features significantly improve registration performance, particularly in the presence of modality-specific artifacts and nonlinear intensity relationships. IMPACT was systematically compared to standard similarity metrics (MSE, NCC, MI, MIND) and several state-of-the-art methods from public challenges. Consistent improvements in TRE, DSC, and HD95 were achieved across all tasks without task-specific training or tuning. By leveraging features from pretrained models like TotalSegmentator and SAM2.1, IMPACT provides a robust alternative to conventional intensity-based metrics and handcrafted descriptors. The plug-and-play integration of these features within both algorithmic and deep learning frameworks ensures broad applicability across diverse datasets and clinical contexts.

In both thoracic CBCT registration tasks (Tasks 1 and 2), strong modality-specific artifacts, such as scatter, truncation, and beam hardening, compromise the reliability of intensity-based metrics, particularly in peripheral and low-contrast regions, limiting the effectiveness of measures like NCC and MI. On Task 1, IMPACT (Jacobian) achieved a median TRE of 1.20 mm [0.85–1.71], outperforming NCC (1.58 mm [1.06–2.92]) and MIND (1.25 mm [0.93–1.85])(Table~\ref{tab:results_task1}). Qualitative results (Fig.\ref{fig:qualitative_task1}) confirm improved anatomical alignment. On Task 2 (Learn2Reg ThoraxCBCT), IMPACT achieved a DSC of 0.652, comparable to Deeds (0.648), which combines MIND with a task-specific discrete optimization framework, and higher than ConvexAdam (0.586), which also uses MIND but with a generic hybrid strategy. This performance secured 6th place out of 21 submissions in the MICCAI-ThoraxCBCT challenge (Table\ref{tab:results_task2}). These results underscore that, while MIND provides a strong handcrafted baseline, its performance is sensitive to the optimization framework, whereas IMPACT offers generalizable accuracy without task-specific engineering. 

In the Lung CT registration task (Task 5), characterized by substantial anatomical variability and large respiratory deformations, IMPACT combined with Elastix achieved competitive performance, ranking third on the final leaderboard of the MICCAI-LungCT challenge with a mean TRE of 1.89 mm (Table \ref{tab:results_task5}). It ranked ahead of the SAME method, which is specifically designed for monomodal registration. A detailed test set analysis (Table~\ref{tab:ablation_task5}) showed that the best configuration achieved a TRE of $1.32 \pm 1.93$ mm, outperforming MIND ($1.75 \pm 3.18$ mm) and all intensity-based baselines.

The observed improvements on these 3 tasks are largely attributable to the early layers of pretrained segmentation networks, which act as learned feature denoisers. These layers effectively suppress high-frequency noise and modality-specific artifacts, while preserving and enhancing fine anatomical details, such as pulmonary fissures, airway bifurcations, and vascular trees, key landmarks that remain stable despite significant deformations, as shown in Fig.~\ref{fig:featuresMap}. In contrast to intensity-based metrics or handcrafted descriptors, these learned features produce denoised and meaningful anatomical representations, suitable for guiding registration in highly deformable settings.

In the MR/CT tasks (Tasks 3 and 4), IMPACT again demonstrated strong performance, despite substantial intensity differences and anatomical variability between modalities. On the abdomen (Task 3), VoxelMorph combined with IMPACT ranked second out of 26 submissions, outperforming several established methods in both DSC and HD95 (Table~\ref{tab:results_task3}). On the pelvis (Task 4), IMPACT consistently outperformed MI and MIND across all structures (Fig.\ref{fig:boxplot_dice_task4}). The improvement was particularly pronounced for the prostate, where intensity and structural differences between MR and CT are most significant, indicating the method’s ability to align semantically corresponding anatomy despite weak appearance similarity.

These results are consistent with results from the Thorax CBCT/CT tasks and confirm that IMPACT’s structure-alignment capability is not modality-specific. In MR/CT scenarios, performance gains are primarily driven by deep semantic features extracted from the final layers of pretrained segmentation models, which encode high-level anatomical concepts such as organ identity and shape rather than local appearance. These features remain effective despite pronounced differences in intensity and texture between modalities, as anatomical structures remain consistent. By projecting both MR and CT volumes into a shared semantic space, IMPACT enables anatomically meaningful correspondence even in the absence of intensity similarity, something traditional metrics MI and MIND fail to achieve.

The results demonstrate that IMPACT consistently matches or surpasses learned and handcrafted methods such as ConvexAdam, SAME, and DEEDS on challenging public registration benchmarks, all while using the widely adopted Elastix framework and requiring no training or task-specific tuning.


As shown in Fig.\ref{fig:boxplot_TRE_task1}, IMPACT in Jacobian mode achieves the highest final accuracy and significantly faster convergence than both static mode and traditional intensity-based metrics. By 200 iterations, Jacobian mode reaches a low median TRE with reduced variability, while intensity-based methods converge more slowly and remain less accurate. Although static mode improves over traditional metrics, its convergence is slower and more variable. These results show that Jacobian optimization provides more informative gradients, accelerating convergence and stabilizing alignment early. However, Jacobian mode is more computationally demanding, especially with deep feature extractors like SAM (Table\ref{tab:computational_cost}). Static mode is more efficient but depends strongly on the spatial resolution of the feature maps, becoming ineffective with deeper layers or with transformer-based models.


The effectiveness of semantic similarity–based registration depends strongly on the choice of pretrained model used for feature extraction. Models trained for segmentation tasks, as opposed to generic classification, produced more anatomically structured and spatially coherent features. This distinction was consistently reflected in Tasks 1 and 5, where segmentation-based models outperformed classification networks such as ConvNeXt.

Within segmentation models, performance was further enhanced when the pretraining dataset included anatomically relevant structures. For example, M258, trained specifically on CT thoracic anatomy, outperformed the more generic M730 in corresponding tasks, suggesting that anatomical specificity during pretraining improves the relevance of extracted features for alignment.

In contrast, both the architectural design (CNN versus transformer) and the domain on which the models were pretrained had a comparatively limited influence. Despite being trained on natural images, the transformer-based SAM2.1 achieved comparable performance to the CNN-based TotalSegmentator, which was trained on medical images. This finding is significant as it demonstrates that sufficiently expressive architectures can learn semantic representations that are general enough to transfer across domains. Furthermore, ablation studies comparing SAM2.1 and MedSAM fine-tuned on medical images showed identical performance, indicating that domain adaptation had little impact on feature quality. This underscores that the semantic richness introduced by segmentation-based pretraining plays a more decisive role than the pretraining domain itself. These results suggest that models like SAM2.1 can act as truly general-purpose feature extractors, even when applied beyond their original domain.


The ablation studies showed that registration performance is sensitive to the choice of distance metric. Simple magnitude-based measures (L1, L2) consistently outperformed cosine similarity and NCC, especially with deeper features (Tables~\ref{tab:ablation_task1},~\ref{tab:ablation_task5}). These results indicate that the semantic features bring both images into a common representation space, where the absolute differences provide a significant alignment indicator.

The ablation studies further highlighted the impact of feature extraction depth on registration performance. Across tasks, features from the second encoder layer generally yielded superior results (Tables~\ref{tab:ablation_task1}, \ref{tab:ablation_task5}). However, the optimal layer was task-dependent: early encoder layers proved most effective in CT/CBCT registration by suppressing modality-specific artifacts while preserving spatial detail, whereas decoder layers performed better in MR/CT tasks by capturing high-level anatomical semantics necessary to bridge large appearance discrepancies. This behavior is illustrated in Fig.\ref{fig:boxplot_dice_task4}, particularly for complex structures such as the prostate.

This study shows that features from large-scale pretrained segmentation models are effective for multimodal image registration, offering strong generalization across tasks without requiring task-specific training. However, a natural trade-off exists between generality and task specificity, as some anatomically relevant information may not be fully leveraged when features are used as-is, especially since the choice of the pretrained model influences the nature and granularity of the extracted features, potentially limiting performance in highly specialized scenarios. To address this limitation, future work could enhance semantic exploitation by developing adaptive feature-weighting methods or training lightweight embedding networks to model task-specific relationships.

Beyond its performance, IMPACT is also a modular platform that enables systematic experimentation with pretrained models, feature layers, and distance functions. It allows the exploration of feature extraction methods in both algorithmic and deep learning–based registration approaches. This flexibility is especially valuable for challenging modalities like PET or functional MRI, where semantic guidance can help overcome low anatomical contrast or complex intensity variations.

\section{Conclusion}


This study introduces IMPACT, a novel similarity metric for multimodal medical image registration that leverages pretrained segmentation models to guide alignment via high-level anatomical features. By integrating feature representations from TotalSegmentator and SAM-based models, IMPACT addresses key limitations of traditional similarity measures, including sensitivity to noise, artifacts, and modality-specific intensity variations. Integrated into both algorithmic (Elastix) and deep learning-based (VoxelMorph) frameworks, IMPACT consistently improved alignment accuracy over traditional intensity-based and handcrafted metrics across multiple datasets.

While IMPACT demonstrated robust performance, particularly in tasks involving substantial modality variation or artifacts, its effectiveness depends on the choice of feature extractor and distance function. Ablation studies highlighted that pretrained segmentation models, such as TotalSegmentator and SAM2.1, provide superior features for guiding registration, with L1 and L2 losses yielding the most consistent results.

Moreover, based on our conclusions, we propose the pretrained model M730 with 2 layers and L2 features distance measure as the recommended generic feature extractor for IMPACT, balancing robustness and efficiency across different registration tasks. However, if the user wishes to further optimize performance for a specific task, the feature extractors in IMPACT can be replaced with a custom model trained on their own data. This allows for more task-specific features while maintaining the versatility of IMPACT, ensuring seamless integration with frameworks like Elastix and VoxelMorph. This adaptability reinforces IMPACT's role as a scalable and efficient tool, suitable for both research and clinical applications.

The integration of semantic feature alignment into image registration holds considerable promise for clinical applications, including tumor monitoring, surgical planning, and image-guided therapy. The plug-and-play nature of IMPACT, combined with its scalability across multimodal datasets, ensures broad applicability and ease of adoption. Future work should focus on extending generalizability to additional modalities and developing strategies for adaptive feature selection tailored to specific clinical contexts.
 
\section{Code Availability and Integration Details}
The resources for implementing the IMPACT similarity metric, including the PyTorch-based loss function and a version of Elastix with IMPACT, are publicly available for reproducibility and community use at:
\url{https://github.com/vboussot/ImpactLoss.git}. Both resources are maintained under open licenses to support research applications. Detailed explanations of the methodology and its integration into registration pipelines are provided in this article and the associated repository documentation. These tools are designed to enable users to reproduce and adapt the framework to their specific use cases, ensuring both accessibility and extensibility.

\section*{Acknowledgment}
The work presented in this article was supported by the Brittany Region through its Allocations de Recherche Doctorale framework and by the French National Research Agency as part of the VATSop project (ANR-20-CE19-0015). Additionally, it was supported by a PhD scholarship Grant from Elekta AB (C.Hémon). The authors have no relevant financial or non-financial interests to disclose. While preparing this work, the authors used ChatGPT to enhance the writing structure and refine grammar. After using these tools, the authors reviewed and edited the content as needed and took full responsibility for the publication’s content.

\bibliographystyle{elsarticle-num}
\bibliography{Biblio.bib}

\onecolumn
\appendix
\section*{Pretrained Models Utilized} \label{sec:pretrainedmodel}
\begin{table*}[h!]
\caption{List of pretrained models used in this study, covering a range of feature extraction paradigms, from handcrafted descriptors (MIND) and 2D foundation models (SAM2.1, MedSAM2) to classification networks (ConvNeXt) and segmentation models pretrained on medical datasets. The field of view indicates the spatial receptive field at the selected extraction layer.}
\centering
\footnotesize
\renewcommand{\arraystretch}{1.2}
\resizebox{1\textwidth}{!}{%
\begin{tabular}{|l|p{12cm}|c|}
\hline
\textbf{Model} & \textbf{Specialization} & \textbf{Field of View} \\ 
\hline
\texttt{MIND} & A handcrafted model designed to extract MIND descriptors \cite{heinrich_mind_2012}. It is parameterized by a radius $r$ and a dilation $d$. & $2r d + 1$ \\ 
\hline
\texttt{SAM2.1} & Segment Anything Model (SAM) 2.1 \cite{kirillov2023segment}, a 2D foundation model designed for general segmentation tasks. & $29$\\ 
\hline
\texttt{MedSAM2} & A medical adaptation of SAM \cite{ma2024segment}, specifically fine-tuned for medical image segmentation. & $29$\\ 
\hline
\texttt{ConvNeXt} & ConvNeXt Tiny \cite{liu2022convnet}, a 2D model trained for a classification task on natural images. & $13$\\ 
\hline
\texttt{M258} & Specialized in lung vessel segmentation. &\multirow{10}{*}{\begin{minipage}{1.5cm}
    $2^l+3$ where $l$ is the layer number
\end{minipage}}\\ 
\cline{1-2}
\texttt{M291} & Focused on organ segmentation, particularly structures such as lung lobes and the liver. &\\ 
\cline{1-2}
\texttt{M292} & Designed for vertebrae segmentation.& \\ 
\cline{1-2}
\texttt{M293} & Tailored for cardio-respiratory system segmentation, including structures such as the aorta and myocardium. &\\ 
\cline{1-2}
\texttt{M294} & Optimized for muscle segmentation, including brain structures.& \\ 
\cline{1-2}
\texttt{M295} & Dedicated to rib segmentation.& \\ 
\cline{1-2}
\texttt{M297} & CT model trained at a 3mm resolution& \\ 
\cline{1-2}
\texttt{M298} & CT model trained at a 6mm resolution& \\ 
\cline{1-2}
\texttt{M730} & Designed for organ segmentation in MRI and CT imaging.& \\ 
\cline{1-2}
\texttt{M731} & Specialized in muscle segmentation in MRI and CT. &\\ 
\cline{1-2}
\texttt{M732} & MRI and CT model trained at a 3mm resolution.& \\ 
\cline{1-2}
\texttt{M733} & MRI and CT model trained at a 6mm resolution.& \\ 
\hline
\end{tabular}
}
\label{tab:models}
\end{table*}

\subsection{TotalSegmentator Models}  
All TotalSegmentator models share a common autoencoder architecture with skip connections. The downsampling process is performed using stride convolutions (stride = 2), while upsampling is achieved through transposed convolutions. Each resolution level consists of two convolutional blocks, with feature dimensions progressively increasing as [32, 64, 128, 320, 320]. Each block comprises a convolutional layer, followed by instance normalization and a Leaky ReLU activation function. Layers are extracted at each level.\\

Preprocessing for TotalSegmentator models:
\begin{itemize}
    \item CT images: Standardized with a canonical orientation, intensities clipped to the range $[-1024, 276]$, and normalized to a mean of $-370$ and a variance of $436$.
    \item MRI images: Standardized with canonical orientation and intensity standardization.
\end{itemize}

\subsection{SAM2.1 and MedSAM2}  
The encoders of SAM2.1 and MedSAM2 are built upon Hiera, a hierarchical vision transformer architecture designed for efficient multi-scale feature extraction \cite{ryali2023hiera}.

Hiera is a hierarchical transformer that refines self-attention across multiple scales. It processes images as non-overlapping patches, extracts multi-scale features using transformer blocks, and progressively reduces spatial resolution while enriching feature representations. Unlike standard vision transformers, Hiera applies local attention at lower levels and maintains global context at higher levels, improving efficiency. Skip connections help merge features across scales, preserving spatial details while enabling hierarchical abstraction. Features are extracted at each downsampling stage.\\

Preprocessing for models pretrained on natural images.  
\begin{itemize}
    \item SAM2.1, MedSAM2 and ConvNeXt utilize ImageNet normalization, where image intensities are standardized using a mean and variance derived from ImageNet statistics to ensure consistency in feature representation.
\end{itemize}

\section{Hyperparameters of the IMPACT Similarity Metric}

Table \ref{tab:LossParameters} lists the hyperparameters available for configuring the IMPACT metric. These settings govern the sampling strategy, multi-resolution scheme, feature extraction, and distance computation. Unless stated otherwise, the same parameters were used for both fixed and moving images.

\begin{table*}[h!]
\caption{Hyperparameters for the IMPACT metric used in Elastix-based registration. Parameters cover spatial sampling, feature extraction, and similarity computation.}
\centering
\footnotesize
\renewcommand{\arraystretch}{1.2}
\resizebox{\textwidth}{!}{%
\begin{tabular}{|l|l|c|}
\hline
\textbf{Parameter} & \textbf{Description} & \textbf{Default value}\\ \hline
\texttt{MaximumNumberOfIterations} & Number of iterations for the optimization process at each resolution. & 500\\ \hline
\texttt{NumberOfSpatialSamples} & Number of spatial samples used during optimization. & 2000\\ \hline
\texttt{NumberOfResolutions} & Number of multi-resolution levels used in the registration process. & 3\\ \hline
\texttt{FinalGridSpacingInPhysicalUnits} & Grid spacing for the final resolution in physical units. & 8\\ \hline
\texttt{ModelsPath} & Path to the pretrained models used for feature extraction. & "Path"\\ \hline
\texttt{Dimension} & Dimensionality of the input images used by the model (2D or 3D). & 3\\ \hline
\texttt{NumberOfChannels} & Number of input channels in the model (like grayscale = 1, RGB = 3).& 1 \\ \hline
\texttt{PatchSize} & Size of the patches extracted for processing. & 5*5*5\\ \hline
\texttt{VoxelSize} & Resampled voxel size for input images. & 1.5*1.5*1.5 \\ \hline
\texttt{LayersMask} & Binary mask indicating which feature extractor layers are selected. & 1\\ \hline
\texttt{SubsetFeatures} & Number of selected feature channels. & 32\\ \hline
\texttt{LayersWeight} & Weight assigned to the selected layers during feature extraction. & 1\\ \hline
\texttt{Mode} & Mode selection for feature extraction (Static, Jacobian). & Jacobian \\ \hline
\texttt{GPU} & GPU configuration (CPU mode (-1) or specific GPU device selection). & -1 \\ \hline
\texttt{FeaturesMapUpdateInterval} & Frequency of feature map updates during optimization. & -1\\ \hline
\texttt{PCA} & Number of principal components retained for the PCA. & 0\\ \hline
\texttt{Loss} & Choose loss comparison for each layer (L1, L2, NCC, Cosine, L1Cosine). & L2\\ \hline
\end{tabular}
}
\label{tab:LossParameters}
\end{table*}

\section*{Choice of Multi-Resolution Strategy with Extracted Features} \label{sec:MultiResolutionStrategy}

\begin{table}[h!]
\caption{Comparison of multi-resolution strategies and their impact on registration accuracy (measured with TRE) on Task 1 after 500 iterations.}
\centering
\footnotesize
\renewcommand{\arraystretch}{1.2}
\resizebox{0.7\textwidth}{!}{%
\begin{tabular}{|l|c|c|}
\hline
\multirow{2}{*}{\textbf{Multi-resolution strategy}} & \multicolumn{2}{|c|}{\textbf{TRE (mm)}}  \\ \cline{2-3}
& Jacobian & Static\\ \hline
Gaussian smoothing + downsampling (standard multi-resolution)& 1.20 & 1.22\\ \hline
Downsampling only (no smoothing) & 1.24 & 1.24\\ \hline
Gaussian smoothing only (no downsampling) & 1.43 & 1.46 \\ \hline
No image pyramid (full-resolution only) & 1.94 & 3.27\\ \hline
\end{tabular}
}
\label{tab:MultiResolutionStrategies}
\end{table}

We conducted an ablation study on Task 1 (Thorax CT/CBCT) to assess the impact of various multi-resolution strategies on registration accuracy using deep features for similarity computation. Table \ref{tab:MultiResolutionStrategies} shows the median TRE after 500 iterations with the second-layer model M258, employing L2 distance for feature similarity computation, across four configurations: downsampling only, Gaussian smoothing only, a combination of both (standard multi-resolution), and no pyramid (full resolution only).

The results show that combining Gaussian smoothing and downsampling yields the most accurate performance, especially in Jacobian mode. Additionally, downsampling alone, without smoothing, still produces competitive results, suggesting that spatial rescaling is sufficient on its own. In contrast, operating at full resolution (without the pyramid) results in significantly poorer performance. These findings highlight the advantage of using a four-level multi-resolution strategy, which combines both smoothing and downsampling, as implemented in all Elastix-based experiments.

\end{document}